\documentclass[10pt,twocolumn,letterpaper]{article}

\usepackage[pagenumbers]{cvpr} 

\definecolor{cvprblue}{rgb}{0.21,0.49,0.74}
\usepackage[pagebackref,breaklinks,colorlinks,allcolors=cvprblue]{hyperref}

\def\confName{CVPR}
\def\confYear{2027}

\author{Parsa Madinei \qquad Ziqi Wen \qquad Miguel Eckstein\\
\textsc{UC Santa Barbara}\\
}

\usepackage[T1]{fontenc}   
\usepackage{booktabs}
\usepackage{multirow}
\usepackage{xspace}
\usepackage{siunitx}
\usepackage{listings}
\usepackage{xcolor}
\usepackage{placeins}   
\graphicspath{{figures/}}
\newcommand{\maybefig}[2][\linewidth]{\IfFileExists{figures/#2.pdf}{\includegraphics[width=#1]{figures/#2.pdf}}{\fbox{\parbox{0.9\linewidth}{\centering\small [figure \detokenize{#2} pending]}}}}
\newcommand{\maybetab}[1]{\IfFileExists{tables/#1.tex}{\input{tables/#1.tex}}{\fbox{\small [table \detokenize{#1} pending]}}}

\newcommand{\apm}{APM\xspace}          
\newcommand{\vpr}{VPM\xspace}          
\newcommand{\pama}{PAMA\xspace}        
\newcommand{\ppma}{PPMA\xspace}        
\newcommand{\ehw}{EHW\xspace}          
\newcommand{\pca}{PJA\xspace}          
\newcommand{\css}{CSS\xspace}
\newcommand{\cssprompt}{``Make your best guess of what might be happening in this scene in one sentence. Avoid
  mentioning objects that do not aid in understanding the context of the scene.''}
\newcommand{\nopeople}{``Do not mention any people.''}
\definecolor{deltaneg}{HTML}{B2182B}
\definecolor{deltapos}{HTML}{1B7837}
\definecolor{attrib}{HTML}{D6604D}
\definecolor{nbA}{HTML}{0F9E8B}
\definecolor{nbB}{HTML}{198BA3}
\definecolor{nbC}{HTML}{2377BB}
\definecolor{nbD}{HTML}{3062C4}
\definecolor{nbE}{HTML}{3F4CBE}
\definecolor{nbF}{HTML}{4E36B8}
\newcommand{\nobodytitle}{\textcolor{nbA}{N}\textcolor{nbB}{O}\textcolor{nbC}{B}\textcolor{nbD}{O}\textcolor{nbE}{D}\textcolor{nbF}{Y}}
\newcommand{\nobodylogo}{\raisebox{-0.42em}[0.82em][0.3em]{\includegraphics[height=1.5em]{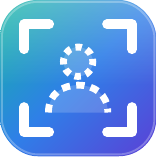}}}

\newcommand{\papertitle}{\nobodylogo\hspace{0.32em}\vspace{0.32em}\nobodytitle{} Was There:\\ Bridging the Human--VLM Gap in Unwarranted Person Attribution Bias}
\newcommand{\dsname}{NOBODY\xspace}
\newcommand{\benchname}{NOBODY-Bench\xspace}

\renewcommand{\dbltopfraction}{0.85}
\renewcommand{\dblfloatpagefraction}{0.8}

\title{\papertitle}

\begin{document}
\twocolumn[{%
\renewcommand\twocolumn[1][]{#1}%
\maketitle
\vspace{-1.7em}
\begin{center}
  \centering
  \captionsetup{type=figure}
  \includegraphics[width=\textwidth]{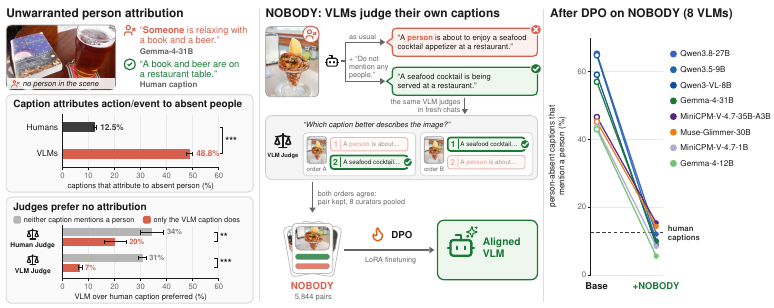}
  \vspace{-1.3em}
  \captionof{figure}{\textbf{Unwarranted person attribution bias and the fix}. \textbf{Left:} For images without people, VLMs attribute events or actions to people much more frequently than human captions do. However, when only the VLM caption includes such an attribution, both human and VLM judges are significantly less likely to prefer it.
  \textbf{Middle:} Each of these VLMs captions a person-free photo twice, in separate chats: as usual and without
  mentioning people. The same VLM then judges its two captions in both orders, and the pairs with agreeing verdicts
  form \dsname, on which we align VLMs through DPO. \textbf{Right:} After alignment on \dsname, the
  eight VLMs attribute absent people about as often as human captions do (dashed line). Error bars show $\pm1$ SE;
  ${}^{**}p<0.01$, ${}^{***}p<0.001$ (bootstrap).}
  \label{fig:teaser}
  \vspace{0.5em}
\end{center}%
}]

\begin{abstract}
    Describing an image requires not only recognizing what is visible but deciding what should be inferred. We investigate a systematic gap between humans and VLMs in their tendency to attribute visible actions or events to people who are not present in an image. We term this behavior \emph{unwarranted person attribution bias} and find that, across 25 SOTA VLMs, captions of person-absent scenes mention a person four times as often as human-written captions of the same images. Through a preference study, we further show that both humans and VLMs prefer captions that avoid such attributions. Motivated by this, we introduce \dsname, a preference dataset designed to bridge the gap between humans and VLMs in unwarranted person attribution. Through direct preference optimization (DPO), we align open VLMs on \dsname and show that their agreement with human captions on person attribution reaches or exceeds inter-human agreement, whereas nine existing public preference datasets leave this behavior largely unchanged. Furthermore, we show that human judges preferred the aligned VLM captions significantly more than those of the baseline model. Alongside \dsname, we release \benchname, which evaluates VLMs for unwarranted person attribution bias and their alignment against human captions and preference. Project page: \href{https://pmadinei.github.io/NOBODY/}{https://pmadinei.github.io/NOBODY/}
    \end{abstract}


\section{Introduction}
\label{sec:intro}

\begin{figure*}[t]
  \centering
  \maybefig[\textwidth]{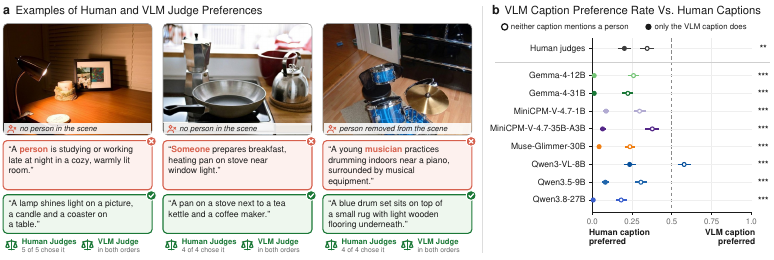}
  \vspace{-1.6em}
  \caption{\textbf{Human and VLM judges prefer the caption without person attribution in person-absent images}.
  \textbf{(a)}~Examples of person-absent images in which only the VLM caption mentions a person (red). Every human
  judge who saw the trial, and the VLM judge (here Qwen3-VL-8B-Instruct) in both presentation orders, chose the human
  caption. The drummer was removed from the third scene. \textbf{(b)}~Share of trials in which the VLM
  caption is chosen over the consensus human caption, for 30 human judges and the eight VLMs that build \dsname. Open
  and filled markers show trials in which neither caption or only the VLM caption has an unwarranted person
  attribution. Error bars show $\pm1$ SE, and stars mark the difference between the two conditions (bootstrap; ${}^{**}p<0.01$,
  ${}^{***}p<0.001$).}
  \label{fig:judges}
\end{figure*}

Vision-language models (VLMs) increasingly write the image descriptions that automated pipelines and agents act on,
so these descriptions should agree with what humans write about the same scenes.

However, VLMs' scene descriptions are not aligned with humans' for scenes without people. VLMs attribute the scene, or an action in it, to a person who is not visible in the image (\cref{fig:teaser},
left). We term this behavior \emph{unwarranted person attribution bias}. Across 22 open VLMs, 48.8\% of the captions
of person-absent scenes mention a person, four times the rate of human captions of the same images (12.5\%).

The human-VLM difference in semantic sensitivity to the presence of people in image captions was first identified by Counterfactual Semantic Saliency (\css)~\cite{wen2026css}, which removes one
object from a scene by inpainting and measures how much the descriptions of the scene change. There, the authors define \emph{Sensitivity to
Person-Non-person}, which is the correlation across removed objects between whether the object is a person and the change that
its removal causes to the semantics of the caption. They show that this sensitivity is lower for VLMs than for human captions, which could arise if VLMs mention people where none are
present~\cite{wen2026css}. We investigate this behavior directly and propose a way to align VLMs with humans on it.

Unwarranted person attribution is not plain object hallucination~\cite{rohrbach2018object,li2023pope}. Human captions
also mention absent people in some cases, so our target is agreement with
human captions, which we measure separately on images with and without people. Existing preference data do not
address this behavior: nine public preference datasets, trained under identical settings, leave it largely unchanged
(\cref{sec:comparison}).

To find a training signal that needs no new human labels, we ran a pairwise preference study. Human judges chose a VLM
caption less often when it was the only caption that attributed a person, and VLMs asked the same question penalized
such captions even more strongly (\cref{fig:judges}). As judges, VLMs thus prefer captions that they do not write
themselves. Motivated by this, we introduce \dsname, a preference dataset in which eight VLMs caption the same
person-free photos as usual and without mentioning people, judge their own two captions in both presentation orders,
and keep the pairs with agreeing verdicts (\cref{fig:teaser}, middle).

Through direct preference optimization (DPO)~\cite{rafailov2023dpo}, we align 22 VLMs on \dsname with shared
training settings. Their agreement with human captions on person-absent images rises for all 22 VLMs, 14 of which
played no role in building \dsname, and reaches or exceeds inter-human agreement for 18 of them, while their
agreement on images with people is not changed significantly (\cref{fig:results}). In follow-up studies, human judges preferred the
aligned captions over the baseline captions, and against human captions the aligned captions were no longer
chosen significantly less often on person-absent images than on images with people. Our contributions are:
\begin{itemize}
  \item We show that VLMs attribute absent people far more often than human captions do, and that human and VLM
  judges both prefer captions without such attributions.
  \item We introduce \dsname, a self-judged preference dataset that raises the agreement of 22 VLMs with human
  captions without affecting their general abilities, and yields captions that human judges prefer.
  \item We release \benchname, which benchmarks the agreement of any VLM to human captions and judges on their unwarranted person attribution behavior when describing scenes.
  \item We investigate the form of these attributions, compare \dsname with nine public preference datasets, study different generations of Proprietary VLMs on these attributions, and ablate our method.
\end{itemize}

\section{Unwarranted Person Attribution}
\label{sec:gap}

\quad\textbf{Images, captions and mentions:} We use the 1,613 images of the \css benchmark~\cite{wen2026css}: 307
natural scenes from OSIE~\cite{xu2014osie} and 1,306 counterfactuals, each with one object removed by inpainting. Each image has about ten human captions (16,603 captions by 227 participants), instructed with the prompt
\cssprompt{} Of these images, 664 are \emph{person-absent}: 562 never contained a person, and in 102 the only person
was removed. Each VLM writes one greedy caption per image with the same prompt. We detect person mentions with a
lexicon (precision 1.00, recall 0.96), and a text-only LLM~\cite{abdin2024phi4} decides
whether a mention \emph{asserts} a present person for our person assertion analysis (see appendix).

\begin{figure*}[t]
  \centering
  \maybefig[\textwidth]{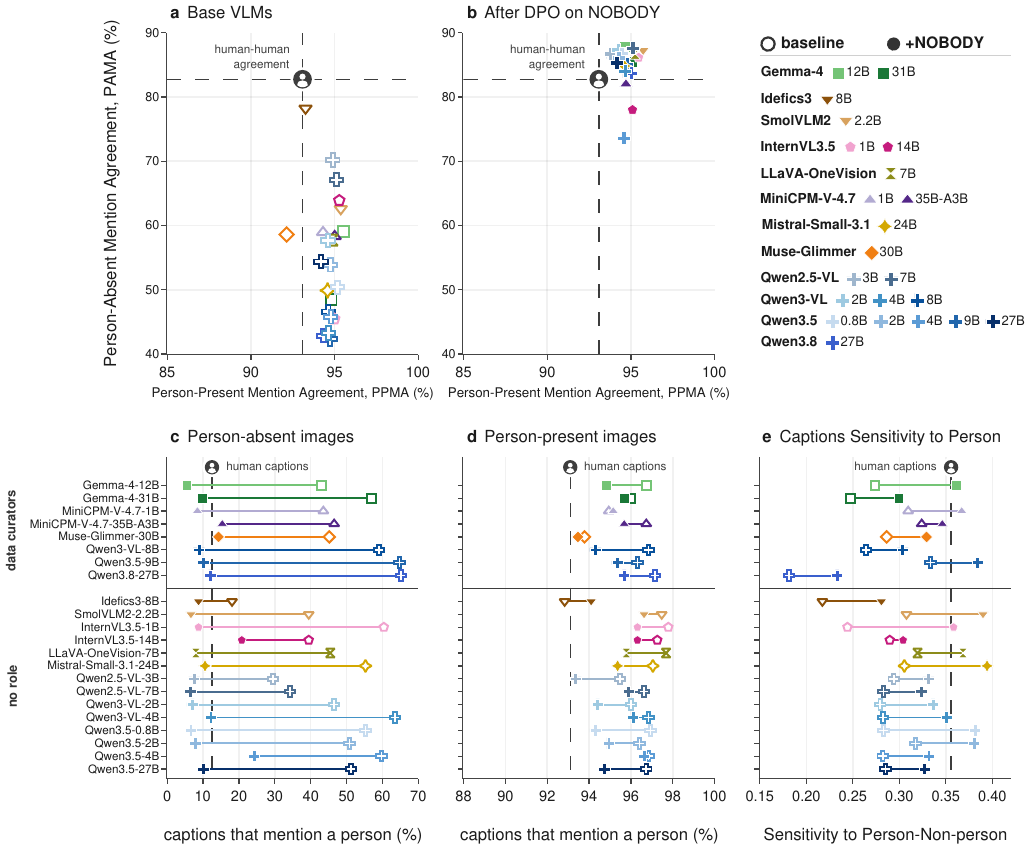}
  \vspace{-1.5em}
  \caption{\textbf{One dataset aligns 22 VLMs with human captions.} \textbf{(a, b)}~Agreement with human captions on
  whether to mention a person, on images with (\ppma, $x$-axis) and without a visible person (\pama, $y$-axis), before
  and after alignment on \dsname. The human icon and the dashed lines mark inter-human agreement
  (each human caption scored against the others). \textbf{(c, d)}~Share of captions
  that mention a person on person-absent (\apm) and person-present images (\vpr), for the baseline (open markers) and
  aligned VLMs (filled markers). Dashed lines mark human captions. \textbf{(e)}~Sensitivity to Person-Non-person of
  the captions in the \css analysis~\cite{wen2026css}. The dashed line marks human captions with all ${\approx}10$
  captions per image; with one caption per image, as for a VLM, the human value is 0.24. Upper rows show the data
  curators, and lower rows the 14 VLMs with no role in building \dsname. Marker shape and hue denote the model
  family.}
  \label{fig:results}
\end{figure*}

\begin{figure*}[t]
  \centering
  \maybefig[\textwidth]{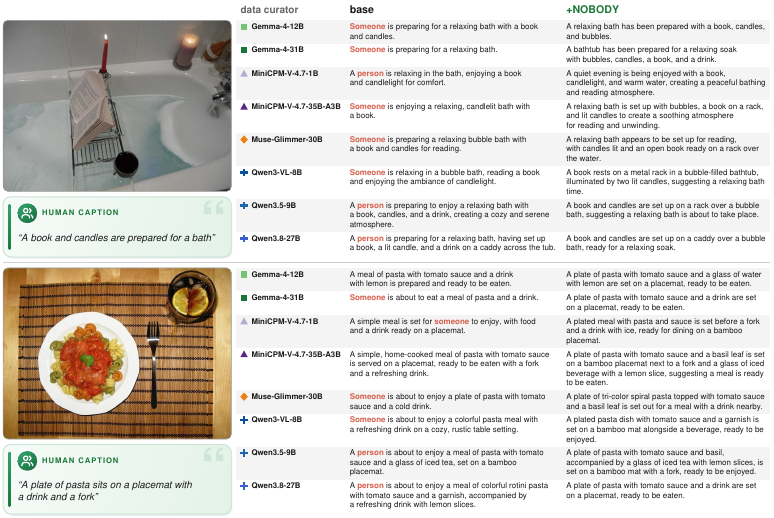}
  \vspace{-1.5em}
  \caption{\textbf{Captions of the eight data curators before and after alignment on \dsname}, on two scenes
  without people. Person words are shown in red, and the consensus human caption appears under each image. More examples are in the appendix.} 
  \label{fig:examples}
\end{figure*}

\textbf{VLMs attributing absent people:} On person-absent images, every VLM exceeds the human rate in attributing a scene to an absent person. This behavior was observed in both scenes that never contained a person and in scenes whose
person was removed (\cref{tab:supp_baseline}). Visible people,
in contrast, are mentioned as often as in human captions (\cref{fig:results}d). Furthermore, we noticed that VLM attributions are far more frequent in scenes with objects that people use (tableware, tools, devices) than with animals or plants, whereas human caption attribution rates do not depend on the objects (see appendix).
This suggests that, rather than describing a hallucinated person in the
picture, the VLM captions usually turn a scene of objects into an action by \emph{someone} (\cref{fig:examples}). 

\textbf{Human and VLM judges attribution preference:} We ran a pairwise preference study in which each trial showed
one \css image and two captions in random order: the consensus human caption of the image, \ie the one most similar
to the image's other human captions, and a greedy caption of Qwen3-VL-8B~\cite{bai2025qwen3vl}. Each participant saw only one version of each scene. Since human and model
judges prefer longer texts~\cite{singhal2024length,park2024length,dubois2024lcalpaca,zheng2023judge}, we constrained
the VLM caption to the length of the human caption, in this study and in both follow-up studies
(\cref{sec:followup}). Human judges clicked the caption that ``better describes the image'' or ``better describes what might be happening in the image''. We controlled for participants' attention by having 10 easy-choice trials, where one caption was completely irrelevant to a scene and another was the correct caption. That gave us 10,670 valid choices from 46 participants.
On person-absent images, human judges chose the VLM caption less often when it was the only caption that attributed a
person than when neither caption mentioned one (\cref{fig:judges}b; mixed-effects odds ratio 0.35, 95\% CI [0.28,
0.44]; 24 of 30 judges). We then asked the eight VLMs that build \dsname the same question in both presentation
orders. They penalize the attribution even more strongly (mean raw odds ratio 0.13 \vs 0.48 for the human judges). 


\section{\dsname: One Dataset to Align Them All}
\label{sec:method}

\quad\textbf{Images:} We draw photos from V3Det~\cite{wang2023v3det}. Unlike COCO~\cite{lin2014coco}, Visual
Genome~\cite{krishna2017visualgenome} or LAION~\cite{schuhmann2022laion}, V3Det feeds none of the public preference
datasets that we compare with, and no image comes from the \css data. A photo counts as person-free only if three
independent checks agree: the V3Det annotations, the OWLv2 detector~\cite{minderer2023owlv2}, and Qwen3-VL-8B asked
whether any person, or part of one, is visible (see appendix).
The curators caption 29,569 such photos and 2,500 photos with a clearly visible person.

\textbf{VLM-judged pairs:} Eight VLMs act as \emph{data curators}. Each curator writes its regular greedy caption of a
photo with the \css prompt. If this caption mentions a person, the curator writes a second caption with the same
prompt followed by \nopeople{} The same VLM then judges its two captions with the same question as the human study, once with each caption first, and we keep the pair only if both orders
give the same answer (consistent verdict). The order matters: Qwen3-VL-8B prefers the rewrite in 85\% of its
judgments when it comes second but in only 57\% when it comes first.
On person-free images, a pair enters \dsname when the rewrite is preferred (chosen) over the regular caption
(rejected). On photos with people, the curator writes both captions for every image, and its consistent verdict is
the label. In three quarters of these pairs, the judges keep the caption with the visible person, which teaches a
model not to drop people who are present.

\textbf{Cleaning, length balancing, and pooling:} First, human judges
penalize captions that assert an absent person more than captions that only imply one (see appendix)
, so we
keep only person-absent pairs whose rejected caption asserts a person and whose chosen caption mentions none. Second,
VLM judges, like human judges, prefer longer captions, and DPO learns this preference: without balancing, captions
grew by up to 44\%. Within each subset and bin of length difference, we therefore keep as many pairs with a longer
chosen caption as with a shorter one. Third, the quality of the judges differs across families: among their
consistent verdicts, the Qwen curators prefer the rewrite in 84.8\% and the MiniCPM curators in 40.3\%, so weak curators yield few pairs. We therefore pool all curators, capping each at 750
person-absent pairs and adding person-present pairs in a 1:3 ratio. \dsname contains 5,844 pairs (4,379
person-absent and 1,465 person-present) on 3,928 images (\cref{tab:supp_dataset}).

\section{Experiments}
\label{sec:experiments}

\subsection{\benchname}
\label{sec:benchmark}

\begin{table*}[t]
  \centering
  \small
  \setlength{\tabcolsep}{4.2pt}
  \caption{\textbf{Comparison with public preference datasets} on Qwen3.5-9B. \apm, \pama and \ppma compare the captions of the \css
  images with human captions, \ehw with the human judges of our study, and \pca with the human judges of four public
  preference datasets. Win is the share of consistent verdicts of an independent VLM judge (Qwen3.5-27B, both orders)
  that prefer the trained model's caption over the baseline caption. Human captions are scored leave-one-out. Bold
  marks the best model.}
  \label{tab:comparison}
  \resizebox{\textwidth}{!}{\begin{tabular}{@{}l l rrrrrrr@{}}
\toprule
 & & \multicolumn{3}{c}{Caption-based Metric} & \multicolumn{2}{c}{Human judges} & VLM judge & \\
\cmidrule(lr){3-5}\cmidrule(lr){6-7}\cmidrule(lr){8-8}
Method & Feedback & APM$\downarrow$ & PAMA$\uparrow$ & PPMA$\uparrow$ & EHW$\uparrow$ & PJA$\uparrow$ & Win$\uparrow$ & Words \\
\midrule
\textit{Human captions} & \textit{leave-one-out} & 12.5 & 82.8 & 93.1 & 43.0 & -- & -- & 12.4 \\
\midrule
Base model & -- & 64.8 & 42.3 & 94.7 & 40.4 & 72.3 & -- & 27.3 \\
Prompting (no absent people) & instruction & 45.2 & 57.9 & 95.4 & 41.7 & 78.2 & 54.4 & 27.1 \\
\midrule
RLHF-V & human corrections & 69.0 & 37.7 & 94.9 & 39.8 & 71.3 & 7.7 & 15.1 \\
POVID & GPT-4V injected hallucinations & 62.5 & 44.2 & \textbf{95.2} & 40.5 & 73.6 & 23.9 & 23.7 \\
VLFeedback (Silkie) & GPT-4V ratings & 64.0 & 42.6 & 94.9 & 40.5 & 72.9 & 44.5 & 29.5 \\
STIC & self + corrupted images & 63.6 & 43.7 & 95.0 & 40.4 & 72.0 & 36.1 & 29.0 \\
CSR & self-reward + CLIP & 64.9 & 42.1 & 95.0 & 40.4 & 72.9 & 39.2 & 27.1 \\
TPR & self, topic-level rewriting & 63.1 & 43.6 & 95.0 & 40.5 & 72.4 & 34.2 & 26.5 \\
HSA-DPO & GPT-4 severity & 61.9 & 44.3 & 95.1 & 40.5 & 74.1 & 23.9 & 24.3 \\
RLAIF-V & open-model feedback & 66.6 & 40.9 & 95.1 & 40.3 & 72.8 & 33.2 & 26.7 \\
HalluVL-DPO & prompt-induced, self & 65.4 & 41.5 & 94.9 & 40.3 & 73.1 & 28.3 & 24.8 \\
\midrule
\textbf{\textcolor{nbA}{N}\textcolor{nbB}{O}\textcolor{nbC}{B}\textcolor{nbD}{O}\textcolor{nbE}{D}\textcolor{nbF}{Y}} & self-judged (8 VLMs) & \textbf{10.1} & \textbf{85.1} & 94.8 & \textbf{43.7} & \textbf{83.6} & \textbf{59.7} & 21.5 \\
\bottomrule
\end{tabular}
}
\end{table*}


\quad\textbf{Agreement with human captions:} A model can lower its rate of person mentions by never mentioning people,
so we measure agreement with human captions separately on images without and with a visible person. Let
$m\in\{0,1\}$ denote whether a caption mentions a person and $h$ the share of the image's human captions that mention
one. The agreement $m\,h+(1-m)(1-h)$ is the probability that a randomly drawn human caption of the image makes the
same decision. Its mean over the 664 person-absent images is the \emph{Person-Absent Mention Agreement} (\pama), and
its mean over the 949 person-present images is the \emph{Person-Present Mention Agreement} (\ppma). For inter-human
agreement, we score each human caption against the other human captions of its image (first row of \cref{tab:comparison}).
The two measures must be read together: a model that never mentions people reaches 87.5\% \pama but only 6.9\%
\ppma. We also report the raw rates Absent-Person Mention (\apm) and Visible-Person mention (\vpr), the shares of person-absent and person-present captions that
mention a person. Taking the majority of an image's human captions as the
correct decision gives the same conclusions (see appendix). 

\textbf{Agreement with human judges:} The \emph{expected human-judge win rate} (\ehw) is the probability that a
human judge prefers a caption over the consensus human caption of the image. We estimate it with a logistic model of
the 10,670 study choices on the differences between the two captions in mention features, with caption length and
object count held at the consensus caption's values so that verbosity is not rewarded (see appendix).
Mentioning an absent person carries the largest weight (odds ratio 0.54). Human captions score 43.0\%, and the model
predicts held-out choices with an AUC of 0.63 (\cref{tab:benchmark_validation}). The \emph{public-judge agreement} (\pca) measures whether a VLM follows human preferences for person attribution across four public preference datasets~\cite{wada2024polos,zhang2025mmrlhf,sun2024llavarlhf,yu2024rlhfv}. Each sample in this benchmark contains an image and two candidate descriptions: one has person attribution, while the other does not. A human judge selected the preferred description. The benchmark includes 3,626 samples and is balanced so that, for half of them, judges preferred the description without person attribution. We compute \pca as the percentage of samples for which the model's inclusion or omission of person attribution agrees with the human preference. A model that always or never has person attribution achieves 50\% agreement.

\begin{figure*}[t]
  \centering
  \maybefig[\textwidth]{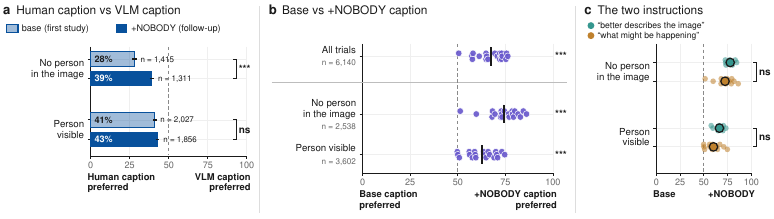}
  \vspace{-1.4em}
  \caption{\textbf{Human judges prefer the aligned captions} (Qwen3-VL-8B, length-matched captions). \textbf{(a)}~Share
  of trials in which the baseline caption or the aligned caption is chosen over the human
  caption, with $\pm1$ SE and a paired image bootstrap test of the difference. \textbf{(b)}~Baseline against
  aligned caption. Each dot is one human judge, ticks mark the mean, $n$ is the number of trials, and stars test the
  judges against 50\% ($t$-test). \textbf{(c)}~The two instructions (``better describes the image'' or ``\dots what
  might be happening in this scene'') by image type, with judges, mean $\pm1$ SE and Mann--Whitney tests.
  ${}^{***}p<0.001$; ns: $p\geq0.05$.}
  \label{fig:followup}
\end{figure*}

\subsection{Setup}
\label{sec:setup}

\quad\textbf{Models:} We align 22 open VLMs (\cref{tab:supp_models}). Eight are the \emph{data curators} whose captions
and judgments make up \dsname: Gemma-4-12B and -31B~\cite{gemma2026gemma4}, MiniCPM-V-4.7-1B and the
mixture-of-experts MiniCPM-V-4.7-35B-A3B~\cite{openbmb2026minicpmv47}, Muse-Glimmer-30B~\cite{meta2026glimmer},
Qwen3-VL-8B~\cite{bai2025qwen3vl}, Qwen3.5-9B~\cite{qwen2026qwen35} and Qwen3.8-27B~\cite{qwen2026qwen38}. The other
fourteen had \emph{no role} in curation, judging or image selection: six come from a curator's family
(Qwen3.5-0.8B, -2B, -4B and -27B, Qwen3-VL-2B and -4B) and eight from other families and developers
(Idefics3-8B~\cite{laurencon2024idefics3}, SmolVLM2-2.2B~\cite{marafioti2025smolvlm}, InternVL3.5-1B and
-14B~\cite{wang2025internvl35}, LLaVA-OneVision-7B~\cite{li2025llavaonevision}, Mistral-Small-3.1-24B~\cite{mistral2025small31},
Qwen2.5-VL-3B and -7B~\cite{bai2025qwen25vl}). All models run in non-thinking mode.

\textbf{Training and evaluation:} We train every model on the same 5,844 pairs with the same settings: DPO with
$\beta=0.1$, LoRA~\cite{hu2022lora} of rank 16 on the language model (vision encoder frozen), learning rate
$5\cdot10^{-6}$ and one epoch. We chose these settings on held-out V3Det photos with two models, never on \css
images, and do not tune them per model. The mixture-of-experts model also adapts its routed experts, at a learning
rate of $2\cdot10^{-5}$. Decoding is greedy and deterministic, and both versions of a VLM share a
backend (see appendix).
We compare means over models with paired $t$-tests; intervals are 95\% scene
bootstraps.

\subsection{\benchname after Alignment}
\label{sec:results}

Alignment on \dsname raises \pama for all 22 VLMs, from 55.3\% to 84.7\% on
average ($p<10^{-11}$), and 18 of them reach or exceed inter-human agreement. \ppma does not change
($p=0.88$; \cref{fig:results}a,b), so the aligned models still describe the people who are
present. The 14 VLMs with no role in building \dsname gain as much as the curators. Their \apm falls from 46.4\% to
10.4\% ($p<10^{-7}$), a relative reduction that does not differ from the curators' (76\% \vs
80\%, Mann--Whitney $p=0.66$). The eight VLMs from families without a curator, which share neither captions nor
judgments with the dataset, fall from 40.2\% to 9.7\% ($p<10^{-3}$). \dsname thus transfers across families,
generations and developers, although its pairs are off-policy for these models. On 400 held-out person-free V3Det
photos, the mean \apm falls from 13.1\% to 2.6\% ($p<10^{-8}$).

The baseline captions describe the scene as an action of \emph{someone},
whereas the aligned captions keep the scene and drop the agent (\cref{fig:examples}). Assertions of an absent person
fall below the human rate (3.4\% \vs 6.5\%), and some removed people become implied agents (``ready to be enjoyed''; see appendix).
\vpr decreases by 1.2 points ($p<10^{-5}$) but stays above the human rate, and caption length
does not change ($+1.3$ words, $p=0.61$). In the \css analysis, the Sensitivity to Person-Non-person rises for all 22
VLMs ($p<10^{-9}$; \cref{fig:results}e), which means that removing a person now changes the captions
more than removing another object. Qwen3.5-4B, whose caption format changes under the shared settings, and
InternVL3.5-14B gain the least. Per-model results and seeds are in the appendix. 

\subsection{Comparison with public preference datasets}
\label{sec:comparison}

We train Qwen3.5-9B, the curator with the largest reduction, on size-matched random samples of nine public preference
datasets, each converted to (image, prompt, chosen, rejected) pairs: RLHF-V~\cite{yu2024rlhfv},
POVID~\cite{zhou2024povid}, VLFeedback~\cite{li2024vlfeedback}, STIC~\cite{deng2024stic}, CSR~\cite{zhou2024csr},
TPR~\cite{he2025tpr}, HSA-DPO~\cite{xiao2025hsadpo}, RLAIF-V~\cite{yu2025rlaifv} and HalluVL-DPO~\cite{halluvldpo2026}.
We also include a prompting baseline (``Do not mention people or animals unless they are visible in the image.'') and
the mDPO objective~\cite{wang2024mdpo} on our data. No public dataset changes \apm or \pama by more than five points
(\cref{tab:comparison}). These datasets target errors about objects and attributes in long answers, and they change
the captions mainly in length: RLHF-V halves it, yet its model attributes people slightly more often. Prompting
removes about 30\% of the attributions. In contrast, DPO on \dsname removes more than four fifths (the mDPO objective
almost as many) and is the best in \apm, \pama, \ehw, \pca and the independent judge's win rate, while \ppma stays
within one point for all methods.

\subsection{General abilities}
\label{sec:general}

\begin{table}[t]
  \centering
  \small
  \setlength{\tabcolsep}{3.5pt}
  \caption{\textbf{General abilities before and after alignment on \dsname.} Accuracy (\%) of the eight curators on
  four benchmarks for the baseline and aligned model. Aligned
  scores are \textcolor{deltaneg}{red} when lower than the baseline and \textcolor{deltapos}{green} when higher. The
  last row gives the mean change in points (paired $t$-test).}
  \label{tab:general}
  \resizebox{\linewidth}{!}{\begin{tabular}{@{}l@{\hspace{5pt}}|@{\hspace{5pt}}rr@{\hspace{5pt}}|@{\hspace{5pt}}rr@{\hspace{5pt}}|@{\hspace{5pt}}rr@{\hspace{5pt}}|@{\hspace{5pt}}rr@{}}
\toprule
 & \multicolumn{2}{c@{\hspace{5pt}}|@{\hspace{5pt}}}{MMStar} & \multicolumn{2}{c@{\hspace{5pt}}|@{\hspace{5pt}}}{MMBench} & \multicolumn{2}{c@{\hspace{5pt}}|@{\hspace{5pt}}}{OCRBench} & \multicolumn{2}{c@{}}{RealWorldQA} \\
\cmidrule(lr){2-3}\cmidrule(lr){4-5}\cmidrule(lr){6-7}\cmidrule(lr){8-9}
Data curator & base & ours & base & ours & base & ours & base & ours \\
\midrule
Gemma-4-12B & 44.3 & 44.3 & 88.0 & \textcolor{deltapos}{88.1} & 78.0 & \textcolor{deltaneg}{77.6} & 66.1 & \textcolor{deltapos}{66.3} \\
Gemma-4-31B & 47.8 & \textcolor{deltaneg}{47.2} & 89.0 & \textcolor{deltaneg}{88.9} & 86.0 & 86.0 & 69.2 & \textcolor{deltaneg}{68.9} \\
MiniCPM-V-4.7-1B & 36.3 & \textcolor{deltaneg}{35.9} & 72.2 & \textcolor{deltaneg}{70.2} & 76.2 & \textcolor{deltapos}{77.6} & 63.8 & \textcolor{deltaneg}{63.4} \\
MiniCPM-V-4.7-35B-A3B & 55.3 & \textcolor{deltapos}{55.4} & 92.0 & 92.0 & 86.7 & 86.7 & 74.9 & \textcolor{deltaneg}{74.2} \\
Muse-Glimmer-30B & 74.3 & \textcolor{deltaneg}{73.5} & 91.0 & 91.0 & 80.6 & \textcolor{deltapos}{81.6} & 76.9 & \textcolor{deltaneg}{76.1} \\
Qwen3-VL-8B & 61.2 & \textcolor{deltapos}{61.6} & 90.0 & 90.0 & 87.2 & \textcolor{deltaneg}{87.1} & 69.5 & \textcolor{deltapos}{69.7} \\
Qwen3.5-9B & 52.5 & 52.5 & 89.7 & \textcolor{deltapos}{89.9} & 86.1 & \textcolor{deltaneg}{85.1} & 74.2 & \textcolor{deltapos}{74.9} \\
Qwen3.8-27B & 50.9 & \textcolor{deltaneg}{50.2} & 90.5 & \textcolor{deltapos}{90.8} & 84.5 & \textcolor{deltapos}{84.6} & 76.3 & \textcolor{deltaneg}{76.1} \\
\midrule
mean change & \multicolumn{2}{c@{\hspace{5pt}}|@{\hspace{5pt}}}{\textcolor{deltaneg}{$-$0.2} ($p=0.17$)} & \multicolumn{2}{c@{\hspace{5pt}}|@{\hspace{5pt}}}{\textcolor{deltaneg}{$-$0.2} ($p=0.48$)} & \multicolumn{2}{c@{\hspace{5pt}}|@{\hspace{5pt}}}{\textcolor{deltapos}{+0.1} ($p=0.65$)} & \multicolumn{2}{c@{}}{\textcolor{deltaneg}{$-$0.2} ($p=0.31$)} \\
\bottomrule
\end{tabular}
}
\end{table}

Alignment on \dsname leaves general abilities intact (\cref{tab:general}). On MMStar~\cite{chen2024mmstar},
MMBench~\cite{liu2024mmbench}, OCRBench~\cite{liu2024ocrbench} and RealWorldQA~\cite{xai2024realworldqa}, evaluated
with VLMEvalKit~\cite{duan2024vlmevalkit}, no mean change across the curators is significant.
POPE~\cite{li2023pope} and CHAIR~\cite{rohrbach2018object} change by about one point or less, and
MMHal-Bench~\cite{sun2024llavarlhf} by one to four of its 96 questions (see appendix). 
An independent judge
(Qwen3.5-27B) prefers the aligned over the baseline caption in 77\% of its consistent verdicts ($p=0.002$).

\subsection{Human judges prefer the aligned captions}
\label{sec:followup}

The preferences in \dsname come from the VLMs themselves, so we test whether human judges agree. We ran two follow-up
studies with Qwen3-VL-8B before and after alignment, with the design, inclusion rule and length matching of the first
study (see appendix).

\textbf{Against human captions:} Eleven human judges chose between the human caption and the aligned caption. On
person-absent images, they chose the human caption less often than against the baseline
caption in the first study ($p<10^{-3}$, paired image bootstrap). On images with people, the two studies do not
differ (\cref{fig:followup}a). The baseline caption was chosen 11.0 points less often on person-absent images than on
images with people ($p<10^{-3}$). For the aligned caption, this extra penalty shrinks to 2.6 points ($p=0.21$).

\textbf{Baseline against aligned:} Twenty human judges chose between the baseline and the aligned caption of an
image, under one of two instructions: ``better describes the image'', as before, or ``better describes what might be
happening in this scene'', which echoes the captioning prompt. Every judge preferred the aligned caption on average
($p<10^{-8}$). The preference is strongest on person-absent images but also holds on
images with people (both $p<10^{-6}$; \cref{fig:followup}b). Within each image type, the two instructions do
not differ (\cref{fig:followup}c). Pooled over image types, the ``what might be happening'' group prefers it
slightly less often (see appendix).

\subsection{Form of the unwarranted attributions}
\label{sec:words}

\begin{figure}[t]
  \centering
  \maybefig[\linewidth]{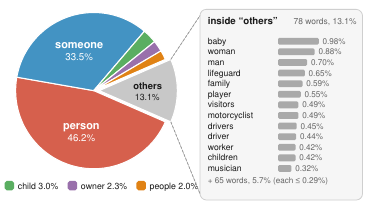}
  \vspace{-1.6em}
  \caption{\textbf{The first person word} of the 7,125 baseline captions of person-absent images that mention a
  person, pooled over the 22 VLMs. Words below 1\% are grouped as ``others'' and listed on the right. Per-model charts
  are in \cref{fig:supp_words}.}
  \label{fig:words}
\end{figure}

Most attributions name nobody in particular. The first person word of a person-absent caption is ``person'' or
``someone'' in four of five captions (\cref{fig:words}; per model in \cref{fig:supp_words}), and specific roles appear
when an object suggests one (``owner'', ``driver'', ``lifeguard''). When a VLM
mentions a person while the majority of the human captions do not, the person is the grammatical subject of an
action in 89\% of the captions (``someone is preparing'', ``a person is enjoying''). When human captions mention a
person in a person-absent image, the person is the subject in only 58\%, and possessives and prepositional phrases
are more common (``a child's room'', ``ready for someone''; see appendix). 
An unwarranted attribution thus has a typical form: a generic agent as the subject of the
activity the objects suggest.

\subsection{Proprietary VLMs}
\label{sec:api}

\begin{figure}[t]
  \centering
  \maybefig[\linewidth]{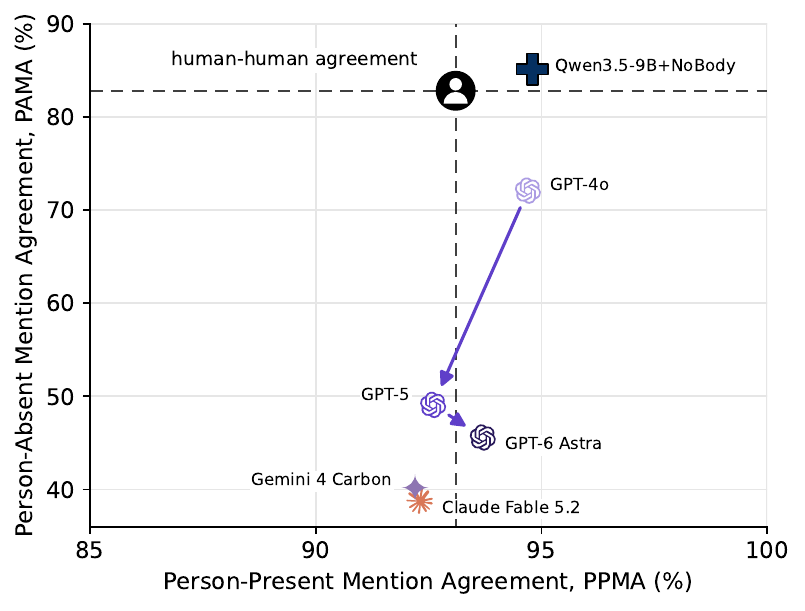}
  \vspace{-1.6em}
  \caption{\textbf{Unwarranted person attribution bias of Proprietary VLMs.} Latest proprietary VLMs also mention a person even in a person-absent scene. With the generational updates of GPT, this bias has become increasingly pronounced.}
  \label{fig:gpt}
\end{figure}


Unwarranted person attribution is not limited to open-source VLMs. We evaluate three closed-source VLMs, GPT-6 Astra, Claude
Fable 5.2 and Gemini 4 Carbon, on \benchname through their APIs, with the prompt and the protocol of the open models
(\cref{fig:gpt}). 
All three mention a person in more than half of their captions of
person-absent images, which places them at the upper end of the 22 open baseline VLMs (\cref{fig:results}c). Their
\pama is far below inter-human agreement, whereas their \ppma is close to it. Since their weights are not available,
we cannot align these models on \dsname.

Newer models do not close this gap on their own. We evaluate three generations of GPT models, GPT-4o, GPT-5 and GPT-6
Astra (\cref{fig:gpt}). With every generation, \pama fall, so each generation is
further from human captions than the one before it. The root cause of this trend needs further investigation. One possible cause is
a larger amount of egocentric visual data in the training of the later generations.

\section{Related Work}
\label{sec:related}

\quad\textbf{Humans and multimodal models:} Studies have addressed diverse alignment challenges in behavioral interpretation~\cite{wang2026subject,kiafar2025mena,benavides2026emotion} and robustness in image reconstruction~\cite{zhou2025deep, saberi2025training}. In visual question answering (VQA), IRIS uses human gaze to resolve ambiguity in questions~\cite{madinei2026iris}. In scene understanding, studies have shown that what people mention depends on the image content~\cite{berg2012importance} and differs from what is present~\cite{misra2016reporting}. \css~\cite{wen2026css} compared human and VLM descriptions through the change caused by removing one object. In this work, we build on \css images and human descriptions.

\textbf{Hallucination in VLM descriptions:} Captioning models mention objects that are not in the
image~\cite{rohrbach2018object}, partly because of co-occurrences in the training text~\cite{biten2022clock,zhou2024lure}.
Benchmarks count hallucinated objects~\cite{rohrbach2018object,wang2023amber}, probe object
existence~\cite{li2023pope} or rate open answers~\cite{sun2024llavarlhf} (see~\cite{bai2024survey}), and mitigation
methods change decoding~\cite{leng2024vcd,huang2024opera,chen2024halc}, revise
outputs~\cite{yin2023woodpecker,zhou2024lure} or tune on curated data~\cite{liu2024lrv}. These benchmarks treat
every absent object as an error, although recent work separates assertive from non-assertive
mentions~\cite{hsu2026mention}. We instead score one behavior against human captions, which also mention absent
people at a nonzero rate.

\textbf{Preference alignment and models as judges:} Preference
optimization~\cite{ouyang2022instructgpt,rafailov2023dpo,azar2024ipo,pang2024rpo,wang2024mdpo} is a standard tool
against hallucination. The preferences come from human corrections~\cite{yu2024rlhfv} or
ratings~\cite{sun2024llavarlhf}, stronger judges~\cite{li2024vlfeedback,xiao2025hsadpo,yu2025rlaifv}, synthetic
rejected answers~\cite{zhou2024povid,zhao2023hadpo,deng2024stic}, or the model's own rewards and
rewrites~\cite{zhou2024csr,he2025tpr,yang2025dsj}, and DPO benefits from on-policy pairs~\cite{yang2025opadpo}.
Models as judges~\cite{zheng2023judge,chen2024mllmjudge,yuan2024selfrewarding} depend on the presentation
order~\cite{zheng2023judge,wang2024fair} and favor longer
answers~\cite{singhal2024length,park2024length,dubois2024lcalpaca} and their own
outputs~\cite{panickssery2024llm}. In \dsname, both captions are the model's own and are judged by the same
model in both orders with balanced lengths, and one pooled dataset also aligns models outside it.

\section{Conclusion}
\label{sec:conclusion}

VLMs attribute scenes without people to someone far more often than human captions do, yet as judges they prefer the
caption without this attribution. Motivated by this, we introduced \dsname, a preference dataset of caption pairs
that VLMs judge consistently. Aligning on \dsname brings 22 VLMs, including models that contributed nothing to it, to
the agreement level of human captions, in captions that human judges prefer. Other behaviors that a model penalizes
as a judge but shows as a writer may be addressed similarly. One possible cause of this bias, which we leave for
future research, is an unbalanced share of egocentric images and videos in the training data of these VLMs.

\paragraph{Limitations.} The preferences in \dsname are the VLMs' own, and VLM judges penalize absent-person attributions more than
human judges do. As a result, this might cause samples which human judges would prefer the caption with such attribution. Furthermore, we evaluate only on English prompts with greedy decoding and automatic person detection.

\section*{LLM Usage Statement}

We used Claude Opus 5.5 and Fable 5.2 to implement experiments according to our detailed step-by-step instructions and confirmed the correctness of the implementation of each step manually. The paper manuscript was written by the authors, which the LLM edited for clarity enhancement and grammatical correctness. We reviewed the code, results, and the text in detail before submission and take full responsibility for the content of this paper. 

{
    \small
    \bibliographystyle{ieeenat_fullname}
    \bibliography{main}
}

\clearpage
\setcounter{page}{1}
\maketitlesupplementary
\appendix
\setcounter{dbltopnumber}{2}
\renewcommand{\dbltopfraction}{0.9}
\renewcommand{\topfraction}{0.9}
\renewcommand{\textfraction}{0.07}
\renewcommand{\dblfloatpagefraction}{0.8}
\renewcommand{\floatpagefraction}{0.8}

\section{First preference study}
\label{sec:supp_study}

Each trial showed one \css image, either an original scene or a counterfactual, together with two
captions placed side by side in random left/right order. One caption was a human caption of that image from the \css
data. The other was written by Qwen3-VL-8B with greedy decoding, using the \css prompt followed by the instruction
``Your response must contain exactly $n$ words.'', where $n$ is the length of the paired human caption. We matched the
lengths because both human and model judges prefer longer
texts~\cite{singhal2024length,park2024length,dubois2024lcalpaca,zheng2023judge}. The judges read the instruction ``You
will see images with two captions. Click the caption that better describes the image.''

We paired the VLM caption with three types of human caption. The \emph{consensus} caption (max\_sim2human) is the
human caption most similar to the other human captions of the image, where similarity is measured in the embedding
space of Jasper~\cite{zhang2024jasper}. The \emph{outlier} caption (min\_sim2human) is the one least similar to the
other human captions. The third type (min\_sim2model) is the human caption least similar to the VLM caption. The pool
contains 4,062 items. Each judge saw each of the 307 scenes once, with one image version and one caption type,
together with ten attention-check trials, all in random order. An attention-check trial paired a caption that
matches the image with a caption of an unrelated image. All analyses in the main text use the consensus caption, which is also the human caption of both follow-up studies.

Of the 62 participants with a valid access code, we included the 46 who saw more than 50 study
images and answered at least 80\% of the attention-check trials correctly (and at least one). These judges contributed
10,670 valid choices, which cover all 307 scenes (one choice per judge and item). Of the excluded participants, five
saw 50 images or fewer, one answered no attention check and ten failed the checks.

\begin{figure*}[t]
  \centering
  \maybefig[\textwidth]{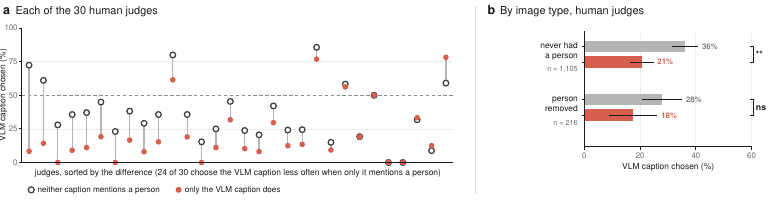}
  \vspace{-1.4em}
  \caption{\textbf{The first study in detail} (consensus human caption, person-absent images). \textbf{(a)}~Share of
  trials in which the VLM caption is chosen, for each human judge who saw both kinds of trial. Open markers show trials
  in which neither caption mentions a person, and filled markers show trials in which only the VLM caption does.
  \textbf{(b)}~The same shares for all human judges, by image type. Error bars show $\pm1$ SE from a bootstrap over
  judges and scenes; ns: $p\geq0.05$.}
  \label{fig:supp_opening}
\end{figure*}

\Cref{tab:supp_judges} compares the human judges with the eight curator VLMs on
person-absent images, using the items with the consensus human caption. For the human judges, we fit a mixed-effects
logistic regression that predicts whether the VLM caption is chosen from whether it is the only caption that mentions
a person, with random intercepts for judge and image and a covariate for counterfactual images. The odds ratio is
0.35 (95\% CI [0.28, 0.44], $z=-9.5$). Of the 30 judges who saw both kinds of trial, 24 chose the VLM caption less
often when it was the only caption that attributed a person (\cref{fig:supp_opening}a). The penalty appears both in
scenes that never contained a person and in scenes whose person was removed, although the sample for the latter is
small (\cref{fig:supp_opening}b).

We further split the trials in which only the VLM caption mentions a
person, using the classifier of \cref{sec:supp_detectors}. Human judges chose the VLM caption in 16.1\% of the 508
trials in which it asserts a present person, but in 32.9\% of the 152 trials in which it only implies one (Fisher's
exact test, $p=2\cdot10^{-5}$). The latter rate is close to the 34.5\% observed when neither caption mentions a person,
which means that implied agents are hardly penalized. For this reason, \dsname targets asserted people.

The three preference studies of this paper (this study and the two follow-up
studies of \cref{sec:supp_followup}) were approved by the Institutional Review Board of the authors' institution. To
preserve anonymity, we withhold the approved protocol number and will state it in the final version. All participants gave informed consent before taking part.

\begin{table}[t]
  \centering
  \small
  \setlength{\tabcolsep}{2.4pt}
  \caption{\textbf{Human and VLM judges} on person-absent images.
  Each VLM judge is asked the study question in a new chat, with greedy decoding and in both presentation orders, and
  only consistent verdicts are counted. The odds ratios are raw odds ratios of choosing the VLM caption when only it
  mentions a person \vs when neither caption does.}
  \label{tab:supp_judges}
  \resizebox{\linewidth}{!}{\begin{tabular}{@{}l rr rr r r@{}}
\toprule
 & \multicolumn{2}{c}{neither mentions a person} & \multicolumn{2}{c}{only the VLM caption does} & & \\
\cmidrule(lr){2-3}\cmidrule(lr){4-5}
judge & $n$ & VLM chosen & $n$ & VLM chosen & odds ratio & order consistency \\
\midrule
human judges & 658 & 34.5\% & 663 & 20.2\% & 0.48 & -- \\
\midrule
Gemma-4-12B & 266 & 25.9\% & 314 & 1.3\% & 0.04 & 92\% \\
Gemma-4-31B & 268 & 22.4\% & 313 & 1.3\% & 0.04 & 93\% \\
MiniCPM-V-4.7-1B & 212 & 29.7\% & 275 & 8.7\% & 0.23 & 78\% \\
MiniCPM-V-4.7-35B-A3B & 228 & 37.7\% & 275 & 6.5\% & 0.12 & 80\% \\
Muse-Glimmer-30B & 266 & 23.7\% & 300 & 4.3\% & 0.15 & 90\% \\
Qwen3-VL-8B & 234 & 57.7\% & 281 & 23.5\% & 0.23 & 82\% \\
Qwen3.5-9B & 252 & 30.6\% & 291 & 8.2\% & 0.20 & 87\% \\
Qwen3.8-27B & 253 & 18.2\% & 308 & 0.6\% & 0.03 & 89\% \\
\bottomrule
\end{tabular}
}
\end{table}



\section{Follow-up preference studies}
\label{sec:supp_followup}

Similar to the first experiment, each trial contained one image
version of each of the 307 CSS scenes, namely the version shown least often so far in the study, and the ten
attention checks of the first study. The caption shown first was balanced, and a returning judge continued where
they had stopped. All VLM captions are greedy captions of Qwen3-VL-8B, before or after alignment on \dsname, with the
length of the consensus human caption of the image, constructed as in the first study.

The first 15 judges compared the human caption with the aligned caption, under the instruction ``Click the caption
that better describes the image.'' Later judges compared the baseline caption with the aligned caption. For these
judges, the instruction alternated by sign-in order between ``Click the caption that better describes the image.''
and ``Click the caption that better describes what might be happening in this scene.'' Image versions whose baseline
and aligned captions are identical (147 of 1,623) were excluded from this second study. The inclusion rule is that of
the first study (more than 50 images and at least 80\% of the attention checks correct). It retains 11 of 13 judges
in the study of human against aligned captions, and 20 of 27 judges (8 and 12 for the two instructions) in the study
of baseline against aligned captions.

\begin{figure*}[t]
  \centering
  \maybefig[\textwidth]{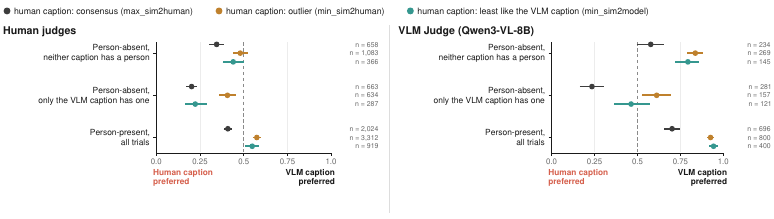}
  \vspace{-1.4em}
  \caption{\textbf{The three human-caption types of the first study.} Share of trials in which the VLM caption
  (Qwen3-VL-8B) is chosen over the consensus human caption, the outlier human caption, or the human caption least
  similar to the VLM caption. The shares are shown by condition for the human judges (left) and for the Qwen3-VL-8B
  judge (right; consistent verdicts only). Error bars show 95\% scene-bootstrap intervals.}
  \label{fig:supp_types}
\end{figure*}

\Cref{fig:followup}a compares the aligned caption (follow-up study) with
the baseline caption (first study) on the same items, using a paired bootstrap over image versions. In the first
study, the judges chose the baseline caption 11.0 points less often on person-absent images than on images with people
($p<10^{-3}$, $t$-test over judges). For the aligned caption, this difference is 2.6 points ($p=0.21$), and the
difference between the two studies is significant (Welch's test, $p=0.018$).

Within each image type, the two instruction groups do not
differ (Mann--Whitney $p=0.24$ on person-absent and $p=0.08$ on person-present images). When we pool over image types,
the ``what might be happening'' group chose the aligned caption in 65.0\% of the trials and the ``describes the
image'' group in 71.0\%. This difference of 6.0 points (95\% CI [0.2, 11.8]) is significant with Welch's $t$-test
($t(16.5)=2.19$, $p=0.044$) but not with the Mann--Whitney test ($p=0.11$). It is small compared with the preference
for the aligned caption, which every judge in both groups shows.

\section{Construction of \dsname}
\label{sec:supp_dataset}

\begin{figure*}[t]
  \centering
  \maybefig[\textwidth]{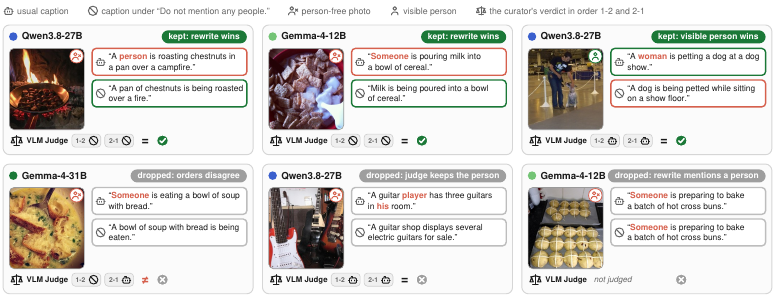}
  \vspace{-1.4em}
  \caption{\textbf{How the pairs of \dsname are selected}, illustrated with six real curation cases. Each card shows
  the curator, the photo (person-free or with a visible person), the curator's usual caption and its caption under the
  instruction \nopeople{} (person words in red), the curator's verdict as a judge in both presentation orders (the icon
  of the winning caption) and the outcome.}
  \label{fig:supp_cases}
\end{figure*}

\paragraph{How pairs are selected.} \Cref{fig:supp_cases} shows six real cases from the curation. Three of them are
pairs that were kept, two on person-free photos and one on a photo with a visible person. The other three are
candidates that were dropped, one for each possible reason: the two presentation orders disagree, the judge keeps the
person on a person-free photo, or the rewrite still mentions a person.

\Cref{tab:supp_dataset} follows each curator from the shared photos to the pairs
that it contributes. Regular captions mention a person far less often on V3Det photos (11.7\% on average) than on \css
scenes without people (48.8\%), because many V3Det photos are close-ups of a single object. Together with the
consistency rule and the cleaning, this rate determines the size of the dataset. The MiniCPM curators, which are weak
judges, keep few pairs. Among its consistent verdicts, MiniCPM-V-4.7-1B prefers its own caption that mentions a person
over its own rewrite in 73\% of the cases, although it penalizes attributions in the pairs of the human study
(\cref{tab:supp_judges}). This suggests that its rewrites, rather than its judgments of people, are the weak link.

\paragraph{Person-present pairs.} For each of the 2,500 photos with a visible person, every curator writes both
captions and judges them, and its consistent verdict becomes the label. In 74.5\% of the person-present pairs of the
dataset, the judges keep the caption that mentions people. In the remaining pairs, the visible people are small or in
the background. As an ablation, we also label these pairs by a rule under which the caption that mentions the visible
people always wins (\cref{tab:supp_ablations}).

\paragraph{Cleaning and balancing.} We drop a person-absent pair when its chosen caption contains a word of the person
lexicon or is classified as asserting a person, or when its rejected caption is not classified as asserting one. We
then balance the caption lengths within each subset (person-absent and person-present) and each bin of absolute
length difference (0, 1--2, 3--5, 6--10 and more than 10 words), keeping as many pairs with a longer chosen caption as
pairs with a shorter one. Each curator contributes at most 750 person-absent pairs and one person-present pair for
every three person-absent pairs. Finally, we remove 28 exact duplicates across curators and hold out 5\% of the images
(299 pairs) for validation. Of the 29,569 person-free photos, 19,569 are multi-object scenes that form the ``first
pool'', and three curators captioned only this pool.


\begin{table*}[t]
  \centering
  \small
  \caption{\textbf{Yield of each curator.} From left to right, the columns give the number of photos captioned, the
  share of regular captions that mention a person (over all photos and over the first pool), the number of pairs
  judged, the share of pairs on which both presentation orders agree, the share of these consistent verdicts that
  prefer the rewrite, the number of person-absent pairs after cleaning, and the number of pairs in \dsname.}
  \label{tab:supp_dataset}
  \resizebox{\textwidth}{!}{\begin{tabular}{@{}l rrrrrr rr@{}}
\toprule
 & photos & \multicolumn{2}{c}{caption mentions a person} & pairs & orders & prefers & after & \multicolumn{1}{c}{in \dsname} \\
Data curator & captioned & all & first pool & judged & agree & rewrite & cleaning & absent + present \\
\midrule
Gemma-4-12B & 29,569 & 5.8\% & 7.5\% & 1,140 & 86\% & 89\% & 731 & 680 + 226 \\
Gemma-4-31B & 29,569 & 8.3\% & 10.7\% & 1,686 & 82\% & 74\% & 792 & 742 + 247 \\
MiniCPM-V-4.7-1B & 29,569 & 8.3\% & 10.6\% & 1,905 & 49\% & 27\% & 117 & 103 + 34 \\
MiniCPM-V-4.7-35B-A3B & 19,569 & 7.9\% & 7.9\% & 983 & 80\% & 56\% & 210 & 122 + 40 \\
Muse-Glimmer-30B & 19,569 & 9.2\% & 9.2\% & 1,400 & 88\% & 77\% & 670 & 509 + 169 \\
Qwen3-VL-8B & 29,569 & 13.7\% & 17.5\% & 3,258 & 70\% & 80\% & 1,437 & 750 + 250 \\
Qwen3.5-9B & 29,569 & 13.4\% & 16.9\% & 2,645 & 85\% & 85\% & 1,327 & 750 + 250 \\
Qwen3.8-27B & 19,569 & 13.3\% & 13.3\% & 1,737 & 83\% & 93\% & 1,066 & 750 + 250 \\
\midrule
\textbf{\dsname} (after removing 28 duplicates) & & & & & & & & 4,379 + 1,465 \\
\bottomrule
\end{tabular}
}
\end{table*}

\section{Detectors and their validation}
\label{sec:supp_detectors}

\paragraph{Person mentions.} Our lexicon extends the person regular expression of the \css analyses in two ways. It
adds role nouns (\eg hunter, official, cameraman, goalkeeper, crew, ``lone figure''), and it adds context rules for
homographs that are frequent in VLM captions (a water \emph{pitcher}, cake \emph{batter}, ``ready to \emph{cook}'', a
music \emph{player}, ceiling \emph{fans}). We validated it on 200 held-out captions that were never used to develop it
(100 from the curation and 100 \css captions of other VLMs). We labelled each caption by hand as asserting, implying or
not mentioning a person and weighted the sample to the population of captions. On this set, the lexicon reaches a
precision of 1.00 and a recall of 0.96 (\cref{tab:benchmark_validation}). The three-way classifier is
phi-4~\cite{abdin2024phi4}, a text-only model that never sees the image. For the class ``asserted'', it reaches a
precision of 1.00 and a recall of 0.91. A caption \emph{asserts} a person if it states or presupposes that someone is
present (``a man is cooking''), including hedged statements (``a person is likely \dots''). A caption \emph{implies} a
person if the person appears only as an owner, a purpose, or a past, generic or hypothetical agent (``ready for
guests'').

\paragraph{Person-free and person-present photos.} V3Det is not exhaustively annotated, so we count a photo as
person-free only if it passes three checks: (i) it has no box of a person category, of a person-implying category or
of a depicted-human category; (ii) OWLv2~\cite{minderer2023owlv2} gives scores below 0.10 for ``a photo of a
person'', ``a human face'' and ``a human hand''; and (iii) Qwen3-VL-8B answers ``no'' to the visibility question of
\cref{sec:supp_prompts}. A photo counts as person-present if it has a person box that covers at least 2\% of the
image and receives a ``yes'' to the real-person question. The held-out V3Det set consists of validation images with
$\text{id} \bmod 3 = 0$, which were never used for curation. It contains 400 person-free photos, 300 photos with a
person box of at least 5\% and 433 photos with object targets.

\section{Details of \benchname}
\label{sec:supp_bench}

\paragraph{Images and references.} The 1,613 \css images comprise 307 OSIE~\cite{xu2014osie} scenes and 1,306
counterfactuals. Of these images, 949 show a person and 664 do not. Among the 664 person-absent images, 562 never
contained a person (105 originals and 457 counterfactuals), and 102 are counterfactuals from which the only person was
removed. Each image has about ten human captions, written with the \css prompt by different participants. We compute
human rates per image and then average them over images, in the same way as for the single caption of a model.

\paragraph{\ehw.} The preference model behind \ehw is a logistic regression of the 10,670 choices of the first study.
Its predictors are the differences between the VLM caption and the human caption in seven features, plus dummy
variables for the type of human caption. The seven features are whether the caption mentions an absent person,
whether it mentions the removed object, whether it mentions the largest, the most central and the most salient
visible object, the number of visible objects that it mentions, and its length. For a new caption, the model predicts
the probability that a judge prefers it over the consensus human caption of the image, with length and number of
objects held at the values of the consensus caption. A caption with the same mentions as the consensus caption scores
43.7\%. Mentioning an absent person has an odds ratio of 0.54, whereas the odds ratios of the other mention features
are close to 1. To obtain the \ehw of human captions, we score every human caption except the consensus one against
the consensus caption, which gives 43.0\%. Cross-validated over held-out judges or held-out images, the model reaches
an AUC of 0.63. Adding a sentence-embedding similarity to the human captions raises the AUC to 0.66, and length and
object count alone reach 0.62.


\begin{table}[t]
\centering
\small
\setlength{\tabcolsep}{4pt}
\caption{Validation of the benchmark. (a) Person-mention detectors on 200 held-out captions with manual labels (precision / recall, population-weighted). (b) Cross-validated AUC of the preference model fitted on the 10,670 human choices (held-out judges / images). (c) Human Sensitivity to Person-Non-person computed from one description per image, the footing of a model's single greedy caption (mean $\pm$ s.d. over 200 draws; with all human captions 0.357).}
\label{tab:benchmark_validation}
\resizebox{\linewidth}{!}{%
\begin{tabular}{l l cc}
\toprule
\multicolumn{2}{l}{(a) detector} & precision & recall \\
\midrule
lexicon v2 & person mentioned (APM, VPM) & 1.00 & 0.96 \\
phi-4 asserted & person asserted (asserted APM) & 1.00 & 0.91 \\
\midrule
\multicolumn{2}{l}{(b) preference model features} & \multicolumn{2}{c}{AUC (judge / image)} \\
\midrule
\multicolumn{2}{l}{mentions only (EHW)} & \multicolumn{2}{c}{0.63 / 0.63} \\
\multicolumn{2}{l}{+ similarity to humans} & \multicolumn{2}{c}{0.66 / 0.66} \\
\multicolumn{2}{l}{length + objects only} & \multicolumn{2}{c}{0.62 / 0.62} \\
\midrule
\multicolumn{2}{l}{(c) human Sensitivity to Person-Non-person, 1 description} & \multicolumn{2}{c}{0.237 $\pm$ 0.021} \\
\bottomrule
\end{tabular}}
\end{table}

\begin{table*}[t]
  \centering
  \small
  \caption{\textbf{Baseline VLMs and human captions} on the \css images (\%). The columns give \apm on the 664
  person-absent images, also split into scenes that never contained a person (562) and scenes whose only person was
  removed (102), followed by the asserted \apm, \vpr, the persistence of the removed object (the share of
  counterfactual captions that still mention the removed non-person object) and the number of words per caption. The
  bottom rows show the open VLMs of the \css paper, which we re-captioned with greedy decoding and did not align.}
  \label{tab:supp_baseline}
  \begin{tabular}{l rrrr rr r}
\toprule
 & \multicolumn{4}{c}{images without a visible person (664)} & & & \\
\cmidrule(lr){2-5}
VLM & \apm & never had one & person removed & asserted & \vpr & persist. & words \\
\midrule
human captions & 12.5 & 12.0 & 15.2 & 6.5 & 93.1 & 14.0 & 12.4 \\
\midrule
\multicolumn{8}{l}{\emph{the 22 VLMs of this paper (base models)}} \\
Gemma-4-12B & 42.9 & 44.5 & 34.3 & 30.9 & 96.7 & 13.2 & 16.2 \\
Gemma-4-31B & 56.9 & 59.4 & 43.1 & 37.3 & 96.0 & 11.7 & 13.1 \\
Idefics3-8B & 18.1 & 18.7 & 14.7 & 11.6 & 92.8 & 12.4 & 12.2 \\
SmolVLM2-2.2B & 39.5 & 40.9 & 31.4 & 37.7 & 97.5 & 17.4 & 14.6 \\
InternVL3.5-1B & 60.4 & 63.2 & 45.1 & 59.3 & 97.8 & 20.1 & 19.6 \\
InternVL3.5-14B & 39.5 & 40.9 & 31.4 & 16.3 & 97.3 & 16.3 & 18.8 \\
LLaVA-OneVision-7B & 45.5 & 46.4 & 40.2 & 29.2 & 97.7 & 17.2 & 16.9 \\
MiniCPM-V-4.7-1B & 43.5 & 43.6 & 43.1 & 19.1 & 94.9 & 15.4 & 21.0 \\
MiniCPM-V-4.7-35B-A3B & 46.5 & 47.5 & 41.2 & 25.9 & 96.7 & 19.2 & 24.2 \\
Mistral-Small-3.1-24B & 55.3 & 56.9 & 46.1 & 27.9 & 97.0 & 9.0 & 16.9 \\
Muse-Glimmer-30B & 45.2 & 46.3 & 39.2 & 33.0 & 93.8 & 13.9 & 19.2 \\
Qwen2.5-VL-3B & 29.5 & 29.4 & 30.4 & 25.2 & 95.5 & 18.5 & 16.6 \\
Qwen2.5-VL-7B & 34.2 & 34.7 & 31.4 & 21.5 & 96.6 & 16.2 & 19.9 \\
Qwen3-VL-2B & 46.5 & 48.9 & 33.3 & 41.7 & 96.0 & 18.0 & 21.3 \\
Qwen3-VL-4B & 63.6 & 64.1 & 60.8 & 45.6 & 96.8 & 16.7 & 24.0 \\
Qwen3-VL-8B & 59.0 & 61.6 & 45.1 & 47.1 & 96.8 & 12.7 & 21.6 \\
Qwen3.5-0.8B & 55.3 & 57.7 & 42.2 & 48.2 & 96.9 & 19.3 & 20.9 \\
Qwen3.5-2B & 50.8 & 53.0 & 38.2 & 42.0 & 96.4 & 21.1 & 23.6 \\
Qwen3.5-4B & 59.8 & 62.1 & 47.1 & 44.1 & 96.8 & 19.9 & 25.9 \\
Qwen3.5-9B & 64.8 & 66.5 & 54.9 & 42.8 & 96.3 & 20.3 & 27.3 \\
Qwen3.5-27B & 51.2 & 52.0 & 47.1 & 30.7 & 96.7 & 19.1 & 23.5 \\
Qwen3.8-27B & 65.2 & 66.4 & 58.8 & 31.9 & 97.2 & 23.4 & 56.7 \\
\midrule
\multicolumn{8}{l}{\emph{open VLMs of the \css paper, re-captioned with greedy decoding}} \\
InternVL3-14B & 33.7 & 34.3 & 30.4 & 17.0 & 96.6 & 16.1 & 17.3 \\
InternVL3-2B & 35.8 & 35.9 & 35.3 & 28.6 & 97.3 & 17.1 & 18.2 \\
InternVL3-8B & 43.2 & 43.6 & 41.2 & 28.9 & 96.5 & 16.6 & 18.8 \\
Qwen3-VL-8B-Think & 53.9 & 55.6 & 44.6 & 45.0 & 94.8 & 12.7 & 15.3 \\
Qwen3-VL-4B-Think & 58.8 & 61.0 & 47.1 & 47.5 & 94.9 & 17.0 & 17.8 \\
Gemma-4-E2B & 59.5 & 61.4 & 49.0 & 56.6 & 95.5 & 9.6 & 13.2 \\
Qwen3-VL-8B & 59.6 & 61.7 & 48.0 & 46.7 & 96.7 & 13.1 & 21.5 \\
Qwen3-VL-4B & 61.4 & 62.1 & 57.8 & 43.8 & 96.9 & 16.4 & 24.0 \\
Gemma-4-E4B & 66.1 & 69.6 & 47.1 & 50.2 & 94.8 & 7.6 & 12.6 \\
\bottomrule
\end{tabular}

\end{table*}

\section{Agreement with the majority of the human captions}
\label{sec:supp_mv}

\begin{figure*}[t]
  \centering
  \maybefig[\textwidth]{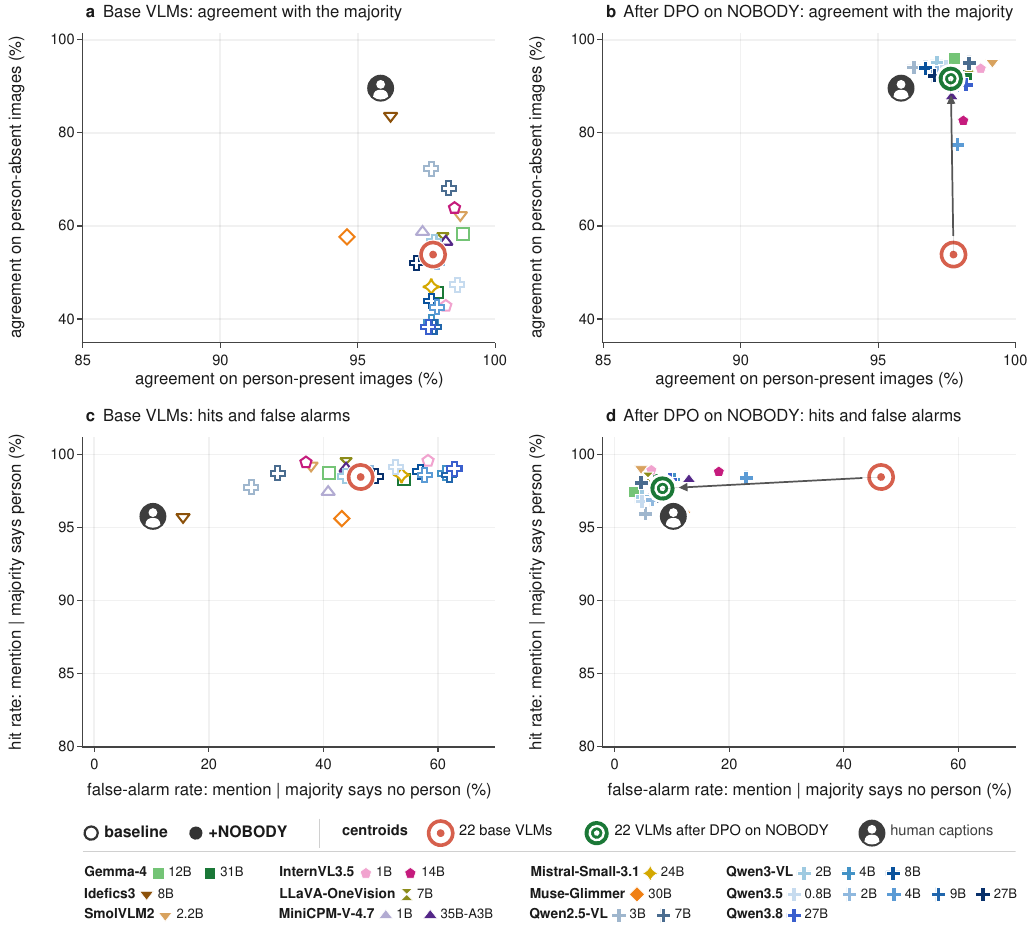}
  \vspace{-1.4em}
  \caption{\textbf{Agreement with the majority of the human captions.} \textbf{(a, b)}~Agreement on person-present
  images ($x$-axis) and on person-absent images ($y$-axis), before and after alignment on \dsname. \textbf{(c,
  d)}~False-alarm rate (the caption mentions a person when the majority mentions none) against hit rate (the caption
  mentions a person when the majority mentions one). Markers show the 22 VLMs, as in \cref{fig:results}. Target badges
  show the centroid of the baseline VLMs (red ring) and of the aligned VLMs (green), and the arrow in (b, d) points
  from the baseline centroid to the aligned centroid. The avatar badge shows the centroid of the 227 participants,
  where the captions of each participant are scored against the majority of the other human captions of every image
  that the participant described.}
  \label{fig:supp_mv}
\end{figure*}

The agreement measures of \cref{sec:benchmark} compare a caption with every human caption of an image. Here, we
instead take the decision of the majority of an image's human captions as correct, and we exclude the 3 person-absent
and 3 person-present images on which the human captions are tied. After alignment, the agreement with the majority
rises on person-absent images from 53.8\% to 91.6\% on average ($p<10^{-11}$; it rises for all 22 VLMs) and does not
change on person-present images (97.7\% and 97.6\%, $p=0.50$; \cref{fig:supp_mv}a,b). We also express this result in
signal-detection terms (\cref{fig:supp_mv}c,d). The false-alarm rate is the share of images on which the caption
mentions a person although the majority of the human captions does not, and the hit rate is the share of images on
which the caption mentions a person when the majority does. The false-alarm rate falls from 46.5\% to 8.3\%
($p<10^{-11}$), while the hit rate falls only from 98.5\% to 97.7\%.

To place individual humans on the same footing, we linked every \css caption to the participant who wrote it, using
the response identifiers of the data collection and matching by exact text. This links 16,604 of the 16,733 responses
and covers all 227 participants. We then scored the captions of each participant against the majority of the
\emph{other} human captions of every image that the participant described (a median of 32 person-absent and 37
person-present images per participant). The centroid of the participants (avatar badge in \cref{fig:supp_mv}) lies at
an agreement of 89.6\% on person-absent and 95.8\% on person-present images, with a false-alarm rate of 10.2\% and a
hit rate of 95.8\%. Finally, we consider the smallest region, trimmed symmetrically on each axis, that holds at least
95\% of the participants (it holds 95.6\%). This region contains all 22 aligned VLMs in both planes. In contrast, 9
baseline VLMs lie below it in agreement on person-absent images, and 10 lie beyond it in false alarms.

\section{\css quantities}
\label{sec:supp_css}

\Cref{tab:supp_main} lists every measure of \benchname for the eight curators, together with two \css quantities
computed from the same greedy captions. The first is Kendall's $\tau$ between the model's and the humans' rankings of
objects by \css~\cite{wen2026css}. The second is the Sensitivity to Person-Non-person, \ie the point-biserial
correlation, across the 1,306 counterfactuals, between whether the removed object is a person and its \css. Both
quantities depend on how many descriptions per image enter \css. Wen \etal~\cite{wen2026css} compare five sampled
model descriptions with about ten human descriptions per image, whereas \benchname has one greedy caption. With one
description per image for both the original and the counterfactual, human captions have a Sensitivity to
Person-Non-person of $0.237\pm0.021$ (200 random draws), compared with 0.357 with all human captions. After alignment
on \dsname, the Sensitivity to Person-Non-person rises for all 22 VLMs, and $\tau$ rises for most of them. We report
these quantities for completeness. The mention measures of \benchname remain the direct measure of the behavior that
we study.

\begin{table*}[t]
  \centering
  \small
  \caption{\textbf{All measures for the eight data curators}, for the baseline model (base) and after alignment on
  \dsname (ours). Values are in \%, except for $\tau\times100$ and the Sensitivity to Person-Non-person. $^\dagger$Human
  value with one description per image, which matches the single caption of a model (0.357 with all human captions).}
  \label{tab:supp_main}
  \resizebox{\textwidth}{!}{\begin{tabular}{@{}l rr rr rr rr rr rr rr rr rr rr@{}}
\toprule
 & \multicolumn{2}{c}{PAMA$\uparrow$} & \multicolumn{2}{c}{PPMA$\uparrow$} & \multicolumn{2}{c}{APM$\downarrow$} & \multicolumn{2}{c}{asserted$\downarrow$} & \multicolumn{2}{c}{VPM} & \multicolumn{2}{c}{persist.} & \multicolumn{2}{c}{EHW$\uparrow$} & \multicolumn{2}{c}{$\tau\uparrow$} & \multicolumn{2}{c}{Sens.\ P--NP} & \multicolumn{2}{c}{PJA$\uparrow$} \\
Data curator & base & ours & base & ours & base & ours & base & ours & base & ours & base & ours & base & ours & base & ours & base & ours & base & ours \\
\midrule
Gemma-4-12B & 59.1 & 87.8 & 95.5 & 94.6 & 42.9 & 5.6 & 30.9 & 2.7 & 96.7 & 94.8 & 13.2 & 17.0 & 41.3 & 43.7 & 42.7 & 48.2 & 0.282 & 0.370 & 78.0 & 83.8 \\
Gemma-4-31B & 48.4 & 85.5 & 94.8 & 95.1 & 56.9 & 9.8 & 37.3 & 2.7 & 96.0 & 95.7 & 11.7 & 16.1 & 40.3 & 43.4 & 41.2 & 46.9 & 0.247 & 0.308 & 75.8 & 83.8 \\
MiniCPM-V-4.7-1B & 58.9 & 85.8 & 94.3 & 94.7 & 43.5 & 8.4 & 19.1 & 2.9 & 94.9 & 95.2 & 15.4 & 18.9 & 41.2 & 43.6 & 46.3 & 48.7 & 0.318 & 0.377 & 76.9 & 82.5 \\
MiniCPM-V-4.7-35B-A3B & 58.3 & 82.0 & 95.0 & 94.7 & 46.5 & 15.4 & 25.9 & 4.4 & 96.7 & 95.7 & 19.2 & 21.7 & 41.4 & 43.4 & 49.4 & 49.2 & 0.323 & 0.369 & 77.2 & 82.5 \\
Muse-Glimmer-30B & 58.6 & 82.0 & 92.1 & 93.0 & 45.2 & 14.3 & 33.0 & 4.7 & 93.8 & 93.5 & 13.9 & 15.4 & 41.0 & 43.0 & 43.6 & 45.6 & 0.297 & 0.336 & 76.1 & 81.8 \\
Qwen3-VL-8B & 46.6 & 86.3 & 94.6 & 94.0 & 59.0 & 8.9 & 47.1 & 3.3 & 96.8 & 94.3 & 12.7 & 19.0 & 40.2 & 43.5 & 40.6 & 46.0 & 0.271 & 0.309 & 70.7 & 82.5 \\
Qwen3.5-9B & 42.3 & 85.1 & 94.7 & 94.8 & 64.8 & 10.1 & 42.8 & 3.5 & 96.3 & 95.4 & 20.3 & 21.6 & 40.4 & 43.7 & 45.0 & 47.6 & 0.341 & 0.389 & 72.3 & 83.6 \\
Qwen3.8-27B & 42.8 & 83.7 & 94.3 & 95.0 & 65.2 & 12.0 & 31.9 & 2.9 & 97.2 & 95.7 & 23.4 & 19.0 & 40.5 & 43.4 & 39.9 & 45.2 & 0.214 & 0.235 & 75.3 & 83.7 \\
\midrule
Human captions & \multicolumn{2}{c}{82.8} & \multicolumn{2}{c}{93.1} & \multicolumn{2}{c}{12.5} & \multicolumn{2}{c}{6.5} & \multicolumn{2}{c}{93.1} & \multicolumn{2}{c}{14.0} & \multicolumn{2}{c}{43.0} & \multicolumn{2}{c}{--} & \multicolumn{2}{c}{0.237$^\dagger$} & \multicolumn{2}{c}{--} \\
\bottomrule
\end{tabular}
}
\end{table*}

\section{Where attributions appear}
\label{sec:supp_triggers}

\begin{figure*}[t]
  \centering
  \maybefig[\textwidth]{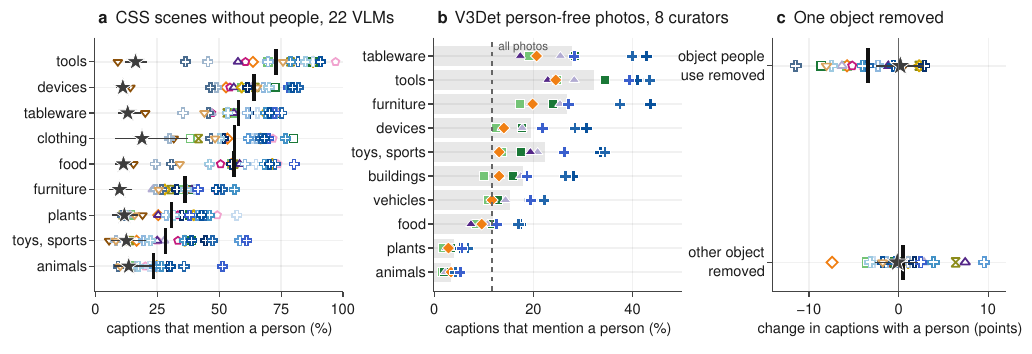}
  \vspace{-1.4em}
  \caption{\textbf{Object domains and unwarranted person attribution.} \textbf{(a)}~Share of captions that mention a
  person in \css scenes that never contained a person, by the object domains in the scene. Open markers show the 22
  baseline VLMs (as in \cref{fig:results}), black ticks show their mean, and stars show the human captions with 95\%
  intervals. \textbf{(b)}~Share of captions that mention a person on 19,569 identical person-free V3Det photos when a
  domain is present, for the eight curators. Bars show the pooled rate, and the dotted line shows the rate over all
  photos. \textbf{(c)}~Change in the share of captions that mention a person when one object is removed from a \css
  scene without people (counterfactual minus original).}
  \label{fig:supp_triggers}
\end{figure*}

We analyze in three ways which objects co-occur with unwarranted attributions (\cref{fig:supp_triggers}). First, we
group the \css scenes that never contained a person by the object domains that they contain. On average, the VLMs
mention a person in 73\% of the scenes with tools, 64\% of the scenes with devices and 58\% of the scenes with
tableware, but in only 23\% of the scenes with animals and 31\% of the scenes with plants. Human captions, in
contrast, average 13.1\% with little variation across domains (SD 2.8; \cref{fig:supp_triggers}a). Second, all eight
curators captioned the same 19,569 person-free V3Det photos. Pooled over the curators, the rate of person mentions is
27.8\% for photos with tableware and 6.4\% for photos without it (a 4.3-fold difference), and the same domains raise
the rate of every curator (\cref{fig:supp_triggers}b). Third, we test the effect of a single object within a scene.
Removing one object that people use from a \css scene that never contained a person lowers the share of person
mentions for 18 of the 22 baseline VLMs (sign test, $p=0.004$), by 3.4 points on average. Removing another object
changes this share in both directions, and human captions do not change in either case (\cref{fig:supp_triggers}c).

\section{When human captions attribute a person}
\label{sec:supp_human_attr}

Human captions of person-absent images also mention a person at a nonzero rate. Here, we examine under which
conditions they do so, and whether the VLMs attribute a person under the same conditions before and after alignment.
As in the preference studies, we represent the human captions of an image by its consensus caption (max\_sim2human).

\paragraph{Conditions of human attributions.} The consensus human caption mentions a person in 39 of the 664
person-absent images (25 scenes). Of these images, 27 never contained an annotated person, and in 12 the only
annotated person was removed. We inspected the 39 images one by one and found four conditions
(\cref{tab:supp_human_attr}), which \cref{tab:supp_human_attr_examples} illustrates with examples. In 16 images, a
person or a body part is in fact visible but is not annotated as a person object of the scene, for example a hand
that holds a cup, a woman who cuts a fish or a man who films with a camera. In 11 images, the person word names a
depicted person (a framed photo, a figurine or the picture on a box) or is part of the name of a non-person (``a
lifeguard station'', ``mother and baby elephant''). In 9 images, the caption infers an owner or a user from
belongings or traces, such as a child's tricycle, the shadow of a dog's owner or an empty chair in front of a running
game. In the remaining 3 images, our lexicon matches a word that refers to an animal or an object (``his leash'' for
a dog), \ie these are detection errors. The consensus human caption thus attributes a person when the image contains
direct evidence of this person: a visible part, a depiction or a belonging. It does not introduce a generic agent
for an activity that the objects merely suggest, which is the typical form of the VLM attributions
(\cref{sec:words}).

\paragraph{VLM attributions under the same conditions.} For each of the 22 VLMs, we compute the share of captions
that mention a person separately on the 39 images whose consensus human caption mentions a person and on the 625
images whose consensus human caption mentions none (\cref{tab:supp_human_attr}). On the 625 images, the baseline VLMs
mention a person in 47.0\% of their captions on average, and the aligned VLMs in 7.1\% ($p<10^{-11}$, paired $t$-test
over VLMs; the rate falls for all 22 VLMs). On the 39 images, the rate falls much less, from 77.5\% to 64.2\%
($p<10^{-7}$; it falls for 21 VLMs). Alignment on \dsname therefore removes most of the attributions that the human
caption does not make and keeps most of the attributions that the human caption also makes. The aligned VLMs keep
the person most often when the person is visible (84.9\%) and least often when an owner or a user must be inferred
from belongings or traces (49.0\%). In the latter case, the aligned caption often describes the belonging without its
owner (\cref{tab:supp_human_attr_examples}). Since the 39 images come from only 25 scenes, we regard the rates of the
individual conditions as descriptive.

\begin{table*}[t]
  \centering
  \small
  \caption{\textbf{Person attributions of human and VLM captions on person-absent images.} The 664 person-absent
  images are split by whether their consensus human caption mentions a person and, if it does, by the condition
  under which it does so. The last two columns give the share of VLM captions that mention a person, averaged over
  the 22 VLMs, before and after alignment on \dsname.}
  \label{tab:supp_human_attr}
  \begin{tabular}{@{}l rr rr@{}}
\toprule
 & & & \multicolumn{2}{c}{VLM captions that mention a person (\%)} \\
\cmidrule(l){4-5}
Consensus human caption & images & scenes & baseline & aligned \\
\midrule
mentions no person & 625 & 193 & 47.0 & 7.1 \\
mentions a person & 39 & 25 & 77.5 & 64.2 \\
\midrule
\quad Person or body part visible, not annotated & 16 & 6 & 96.0 & 84.9 \\
\quad Depicted person or person word for a non-person & 11 & 7 & 73.1 & 63.2 \\
\quad Owner or user inferred from belongings or traces & 9 & 9 & 72.2 & 49.0 \\
\quad Pronoun or noun that refers to an animal or object & 3 & 3 & 10.6 & 3.0 \\
\bottomrule
\end{tabular}

\end{table*}

\begin{table*}[t]
  \centering
  \small
  \renewcommand{\arraystretch}{1.15}
  \caption{\textbf{Examples of person attributions in consensus human captions} of person-absent images, with the
  captions of Qwen3-VL-8B for the same images before and after alignment on \dsname. The captions are verbatim, and
  person words are shown in red.}
  \label{tab:supp_human_attr_examples}
  \begin{tabular}{@{}p{0.15\textwidth}p{0.22\textwidth}p{0.275\textwidth}p{0.275\textwidth}@{}}
  \toprule
  condition & consensus human caption & baseline caption & after alignment on \dsname \\
  \midrule
  \csname @@input\endcsname tables/tab_human_attributions_examples.tex
  \bottomrule
  \end{tabular}
\end{table*}

\section{Person words in unwarranted attributions}
\label{sec:supp_wordsapp}

\begin{figure*}[t]
  \centering
  \maybefig[\textwidth]{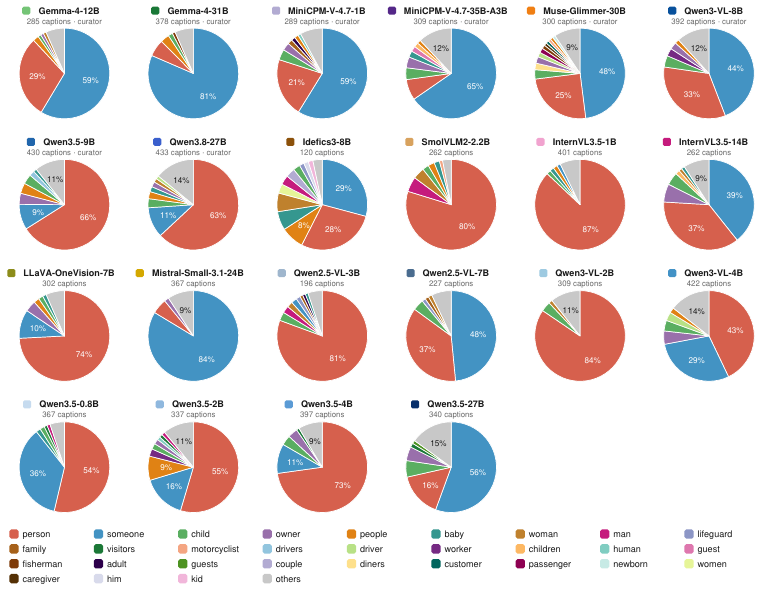}
  \vspace{-1.4em}
  \caption{\textbf{The first person word for each baseline VLM}, in the captions of person-absent \css images that
  mention a person. The data curators are shown first. Words that account for less than 1\% of a model's captions are
  grouped as ``others'' (\cref{tab:supp_others}).}
  \label{fig:supp_words}
\end{figure*}

\Cref{fig:supp_words} shows the first person word for every baseline VLM, and \cref{tab:supp_others} lists the words
that are grouped as ``others'' for each model. The generic words ``person'' and ``someone'' dominate in every model
except Idefics3-8B. This model makes fewer attributions, and they are spread over more specific nouns (``baby'',
``woman'', ``couple'').

\begin{table*}[t]
  \centering
  \scriptsize
  \caption{\textbf{Words grouped as ``others''} in \cref{fig:supp_words}, with the number of captions for each word.}
  \label{tab:supp_others}
  \begin{tabular}{@{}lp{0.80\textwidth}@{}}
\toprule
model & words grouped as ``others'' (each below 1\% of the model's captions; count) \\
\midrule
Gemma-4-12B & family (2), adult (2), man (2), fisherman (1), lifeguard (1), hunter (1), official (1), driver (1), drivers (1), baby (1), musician (1), skydiver (1), pedestrian (1), trainer (1), swimmers (1), passenger (1) \\
Gemma-4-31B & baby (3), children (3), guest (2), drivers (2), musician (2), photographer (1), hunter (1), official (1), driver (1), boy (1), adult (1), owner (1), woman (1), pedestrian (1), visitors (1), motorcyclist (1), camera operator (1), worker (1) \\
MiniCPM-V-4.7-1B & worker (2), visitors (2), player (2), musician (2), audience (2), man (1), surfer (1), customers (1), lifeguard (1), hunter (1), boy (1), drivers (1), guests (1), woman (1), parachutist (1), motorcyclist (1), guest (1), children (1) \\
MiniCPM-V-4.7-35B-A3B & lifeguard (3), motorcyclist (3), individual (3), human (3), mother (2), family (2), drivers (2), guests (2), woman (2), player (2), visitors (2), hunter (1), boy (1), team (1), worker (1), girl (1), viewer (1), diners (1), musician (1), parachutist (1), student (1), cameraman (1) \\
Muse-Glimmer-30B & lifeguard (2), family (2), adult (2), children (2), drivers (2), worker (2), player (2), human (1), guest (1), customers (1), boy (1), traveler (1), woman (1), girl (1), baby (1), musician (1), skydiver (1), performer (1), camera operator (1), swimmers (1), diner (1) \\
Qwen3-VL-8B & human (3), motorcyclist (3), visitors (3), drivers (3), guests (3), player (3), lifeguard (2), family (2), cyclist (2), mother (2), children (2), customers (2), traveler (2), guest (2), musician (2), individual (2), boy (1), viewers (1), newborn (1), diners (1), parachutist (1), rider (1), cameraman (1), driver (1) \\
Qwen3.5-9B & visitors (4), lifeguard (3), couple (3), motorcyclist (3), player (3), family (2), mother (2), woman (2), passersby (2), dancer (1), hunter (1), boy (1), guests (1), traveler (1), customers (1), girl (1), photographer (1), personnel (1), musician (1), performer (1), fisherman (1), audience (1), driver (1), student (1), rider (1), cameraman (1), swimmers (1), owners (1), tourist (1), man (1), worker (1) \\
Qwen3.8-27B & family (4), adult (4), guest (4), fisherman (4), lifeguard (3), man (3), hunter (3), players (3), children (3), player (2), viewer (2), customer (2), fishermen (2), patients (1), customers (1), shepherd (1), individual (1), lady (1), boy (1), traveler (1), girl (1), friends (1), diners (1), groom (1), musician (1), pedestrian (1), visitors (1), rider (1), team (1), visitor (1), swimmers (1), human (1), patient (1), soldier (1), staff (1) \\
Idefics3-8B & girls (1), children (1), lady (1), his (1) \\
SmolVLM2-2.2B & family (1), mother (1), lifeguard (1), boy (1), friends (1), musician (1), player (1), children (1) \\
InternVL3.5-1B & family (4), children (4), couple (3), motorcyclist (3), player (3), man (3), lifeguard (1), pedestrian (1), boy (1), shepherd (1), adult (1), woman (1), musician (1), visitors (1), patient (1) \\
InternVL3.5-14B & family (2), mother (2), lifeguard (2), pedestrian (2), people (2), nun (2), rider (2), guest (1), boy (1), drivers (1), individual (1), friends (1), musician (1), skydiver (1), player (1), visitors (1), motorcyclist (1) \\
LLaVA-OneVision-7B & man (3), players (2), family (2), mother (2), lifeguard (2), boy (1), guests (1), couple (1), customers (1), player (1), musician (1), pedestrian (1), motorcyclist (1), passengers (1) \\
Mistral-Small-3.1-24B & children (3), child (3), player (3), hunter (2), guest (2), rider (2), customer (2), drivers (2), parents (2), motorcyclist (2), mother (1), patient (1), boy (1), traveler (1), baby (1), people (1), diner (1), musician (1), pedestrian (1), swimmers (1), individual (1) \\
Qwen2.5-VL-3B & individual (1), boy (1), drivers (1), people (1), girl (1), pedestrian (1), player (1), driver (1), owner (1), lone figure (1) \\
Qwen2.5-VL-7B & pedestrian (2), owner (2), player (2), people (1), boy (1), shepherd (1), mother (1), viewers (1), baby (1), musician (1), parachutist (1), visitors (1), rider (1) \\
Qwen3-VL-2B & lifeguard (3), man (3), player (3), someone (3), patient (3), baby (3), mother (2), people (2), owner (2), students (1), family (1), hunter (1), boy (1), musician (1), parachutist (1), cameraman (1), swimmers (1), individual (1) \\
Qwen3-VL-4B & human (4), lifeguard (3), patient (3), motorcyclist (3), drivers (3), woman (3), player (3), fisherman (3), family (2), mother (2), worker (2), man (2), visitors (2), customer (2), guest (2), patients (1), tourist (1), customers (1), hunter (1), passenger (1), boy (1), traveler (1), viewers (1), baby (1), viewer (1), friends (1), diner (1), musician (1), parachutist (1), fishermen (1), trainer (1), student (1), rider (1), cameraman (1), passersby (1), swimmers (1), his (1) \\
Qwen3.5-0.8B & lifeguard (2), player (2), people (2), woman (2), family (1), viewers (1), customers (1), girl (1), hunter (1), boy (1), drivers (1), human (1), mother (1), silhouette of a  (1), musician (1) \\
Qwen3.5-2B & lifeguard (3), woman (3), fisherman (3), patient (2), adult (2), viewers (2), player (2), motorcyclist (2), family (1), mother (1), drivers (1), children (1), cyclist (1), boy (1), baby (1), friends (1), diners (1), musician (1), parachutist (1), student (1), passengers (1), audience (1), owners (1), diner (1), cameraman (1), passerby (1) \\
Qwen3.5-4B & lifeguard (3), motorcyclist (3), diner (2), woman (2), baby (2), drivers (1), customers (1), family (1), hunter (1), boy (1), driver (1), adult (1), human (1), patient (1), viewers (1), girl (1), players (1), musician (1), skydiver (1), player (1), audience (1), owners (1), student (1), cameraman (1), passersby (1), tourist (1), man (1) \\
Qwen3.5-27B & lifeguard (3), family (3), motorcyclist (3), customers (3), drivers (3), people (3), woman (3), driver (2), viewers (2), traveler (2), player (2), baby (2), student (2), cyclist (1), adult (1), players (1), worker (1), silhouette of a  (1), hunter (1), boy (1), mother (1), couple (1), girl (1), musician (1), performer (1), fisherman (1), rider (1), visitor (1), diner (1), cameraman (1), surfer (1) \\
\bottomrule
\end{tabular}

\end{table*}

\section{Syntactic role of the attributed person}
\label{sec:supp_syntax}

\begin{figure}[t]
  \centering
  \maybefig[\linewidth]{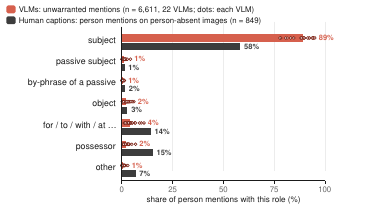}
  \vspace{-1.4em}
  \caption{\textbf{Syntactic role of the first person word} in VLM attributions and in human captions. The VLM
  attributions are the person mentions on person-absent images that the majority of the human captions does not share
  (dots show the individual VLMs). The human captions are those of person-absent images that mention a person.}
  \label{fig:supp_syntax}
\end{figure}

We analyze the syntactic role of the person with the dependency parser of spaCy (\texttt{en\_core\_web\_trf}). We
parsed every baseline caption of a person-absent image in which a VLM mentions a person while the majority of the
image's human captions do not (6,611 captions of 22 VLMs). For reference, we also parsed the 849 human captions of
person-absent images that mention a person. We read the role of the first person word from the parse and distinguish
six roles: nominal subject (including the head of a noun phrase modified by a participle clause, as in ``a person
enjoying a drink''), subject of a passive, \emph{by}-phrase of a passive, object, object of a preposition, and
possessor. Conjuncts, appositions and ``a group of'' inherit the role of their head. In a manual check of 60 random
parses, the role was correct in 58, and the two errors were false positives of the lexicon (``four individual
servings'', ``a woman figurine''). In 89\% of the VLM attributions, the person is the subject of an active clause
(\cref{fig:supp_syntax}), most often of the verbs \emph{prepare}, \emph{enjoy}, \emph{work}, \emph{take} or
\emph{sit}. In human captions of person-absent images, the mentioned person is the subject in only 58\% of the cases
and is much more often a possessor (15\%, ``a child's room'') or the object of a preposition (14\%, ``ready for
someone'').

\section{Implementation details and compute}
\label{sec:supp_impl}

\paragraph{Models and inference.} \Cref{tab:supp_models} lists the models with their Hugging Face repositories. We
follow each model card for loading and chat templates, and we switch thinking off. Muse-Glimmer-30B cannot disable its
reasoning, so we run it at the reasoning strength ``low'' and keep only its final answer. For the two model families
whose default system prompt contains the current date (Muse-Glimmer and Mistral-Small), we pass the prompt with a
fixed date. We scale the images so that their longer side is at most 1,195 pixels, which is the width of the \css
stimuli. Before scoring, we remove a leading meta-preamble that ends in a colon (``Based on the image provided, here
is \dots:''), in the same way for baseline and aligned captions.

\paragraph{Proprietary VLMs.} We queried GPT-6 Astra (\texttt{gpt-6-astra}), Claude Fable 5.2
(\texttt{claude-fable-5-2}) and Gemini 4 Carbon (\texttt{gemini-4-carbon}) through the APIs of their developers in
November 2026. As for the open models, each caption comes from a new chat that contains only the image and the \css
prompt, without a system prompt, the longer image side is at most 1,195 pixels, and the answer is limited to 400
tokens. We requested the setting closest to greedy decoding without thinking that each API accepts. Gemini 4 Carbon
ran at temperature 0 with thinking switched off. GPT-6 Astra and Claude Fable 5.2 accept neither a temperature nor
switching their reasoning off, so they ran with their default sampling at the lowest reasoning effort (``low'').
Their captions are therefore single samples and are not deterministic. GPT-4o (\texttt{gpt-4o}) has no reasoning setting and ran with its default sampling. We scored the captions of the 1,613 \css
images and of the 3,626 images of the public preference sets (\cref{sec:supp_comparison}).

\paragraph{Determinism.} Greedy decoding alone is not reproducible on GPUs, because the results depend on the
composition of the batch. For the Qwen3-VL, Gemma-4, Muse-Glimmer, InternVL3.5, SmolVLM2, Qwen2.5-VL and Idefics3
models, we use vLLM~\cite{kwon2023vllm} with batch-invariant kernels, which reproduces 100\% of the captions across
repeated calls, batch compositions and a batch size of 1. The hybrid Gated-DeltaNet models (Qwen3.5, Qwen3.8 and
MiniCPM-V-4.7) have no batch-invariant kernel in vLLM. We therefore evaluate them with
transformers~\cite{wolf2020transformers}, using fixed batches, a fixed order and deterministic algorithms. We
evaluate LLaVA-OneVision-7B and Mistral-Small-3.1-24B in the same way, because vLLM does not support LoRA for the
former and does not apply the tokenizer fix of the latter. The baseline and aligned captions of a model always come
from the same backend.

\paragraph{Training.} We apply LoRA~\cite{hu2022lora} (rank 16, $\alpha=32$, dropout 0.05) to the attention and MLP
projections of the language model and, for the hybrid models, also to the linear-attention projections. The vision
encoder and the projector are frozen. We use the sigmoid DPO loss with $\beta=0.1$ and AdamW without weight decay.
The learning rate is $5\cdot10^{-6}$, with 10\% linear warm-up and cosine decay, and each step uses 32 pairs
(micro-batches of 4). We further use gradient checkpointing and bf16 precision, and we precompute the reference
log-probabilities with the adapters disabled. One epoch over \dsname amounts to 174 optimizer steps. On one shared
GPU, it took between 20 minutes (Qwen3.5-0.8B) and about ten hours (LLaVA-OneVision-7B in transformers). The runs
that train a curator on its own data only use two epochs, since these sets are about four times smaller. For the
mixture-of-experts MiniCPM-V-4.7-35B-A3B we also apply LoRA to the routed experts and use a learning rate of
$2\cdot10^{-5}$; data and all other settings are unchanged.

\paragraph{Selection of the training settings.} We swept $\beta\in\{0.1,0.3\}$ and the learning rates
$\{5\cdot10^{-6},10^{-5}\}$ on MiniCPM-V-4.7-1B and Qwen3.5-9B, using the length-balanced pairs of the first curator
(1,469 pairs, two epochs). We evaluated each configuration on the held-out V3Det photos and never on \css images. A
configuration passes if \vpr drops by at most two points, the caption length changes by at most 15\% and the share
of one-sentence captions drops by at most five points. The configuration with $\beta=0.1$ and a learning rate of
$5\cdot10^{-6}$ is the only one that passes on both models (\cref{tab:supp_sweep}).


\paragraph{Compute.} All experiments ran on two NVIDIA RTX PRO 6000 Blackwell GPUs (96\,GB) with 32 CPU cores, which
we shared with another user. In total, they took roughly 250 GPU-hours, of which about 160 were spent on training, 60
on captioning and evaluation, 11 on curation and 15 on the general benchmarks and judging.

\begin{table*}[t]
  \centering
  \small
  \caption{\textbf{Models} used in this paper, with their role, their size and the backend used for evaluation
  (\cref{sec:supp_impl}).}
  \label{tab:supp_models}
  \resizebox{\textwidth}{!}{\begin{tabular}{l l l r l}
\toprule
VLM & Hugging Face repository & role & params & eval.\ backend \\
\midrule
Gemma-4-12B & \texttt{google/gemma-4-12B-it} & data curator & 12B & vLLM \\
Gemma-4-31B & \texttt{google/gemma-4-31B-it} & data curator & 31B & vLLM \\
MiniCPM-V-4.7-1B & \texttt{openbmb/MiniCPM-V-4.7-1B} & data curator & 1.55B & transformers \\
MiniCPM-V-4.7-35B-A3B & \texttt{openbmb/MiniCPM-V-4.7-35B-A3B} & data curator & 35.2B (3.5B active) & transformers \\
Muse-Glimmer-30B & \texttt{meta-models/Muse-Glimmer-30B} & data curator & 29.6B & vLLM \\
Qwen3-VL-8B & \texttt{Qwen/Qwen3-VL-8B-Instruct} & data curator & 8B & vLLM \\
Qwen3.5-9B & \texttt{Qwen/Qwen3.5-9B} & data curator & 9B & transformers \\
Qwen3.8-27B & \texttt{Qwen/Qwen3.8-27B} & data curator & 27B & transformers \\
Idefics3-8B & \texttt{HuggingFaceM4/Idefics3-8B-Llama3} & no role & 8.5B & vLLM \\
SmolVLM2-2.2B & \texttt{HuggingFaceTB/SmolVLM2-2.2B-Instruct} & no role & 2.2B & vLLM \\
InternVL3.5-1B & \texttt{OpenGVLab/InternVL3\_5-1B-HF} & no role & 1.1B & vLLM \\
InternVL3.5-14B & \texttt{OpenGVLab/InternVL3\_5-14B-HF} & no role & 15.1B & vLLM \\
LLaVA-OneVision-7B & \texttt{llava-hf/llava-onevision-qwen2-7b-ov-hf} & no role & 8B & transformers \\
Mistral-Small-3.1-24B & \texttt{mistralai/Mistral-Small-3.1-24B-Instruct-2503} & no role & 24B & transformers \\
Qwen2.5-VL-3B & \texttt{Qwen/Qwen2.5-VL-3B-Instruct} & no role & 3.8B & vLLM \\
Qwen2.5-VL-7B & \texttt{Qwen/Qwen2.5-VL-7B-Instruct} & no role & 8.3B & vLLM \\
Qwen3-VL-2B & \texttt{Qwen/Qwen3-VL-2B-Instruct} & no role & 2.1B & vLLM \\
Qwen3-VL-4B & \texttt{Qwen/Qwen3-VL-4B-Instruct} & no role & 4.4B & vLLM \\
Qwen3.5-0.8B & \texttt{Qwen/Qwen3.5-0.8B} & no role & 0.8B & transformers \\
Qwen3.5-2B & \texttt{Qwen/Qwen3.5-2B} & no role & 2B & transformers \\
Qwen3.5-4B & \texttt{Qwen/Qwen3.5-4B} & no role & 4B & transformers \\
Qwen3.5-27B & \texttt{Qwen/Qwen3.5-27B} & no role & 27B & transformers \\
\bottomrule
\end{tabular}
}
\end{table*}

\begin{table}[t]
  \centering
  \small
  \setlength{\tabcolsep}{3pt}
  \caption{\textbf{Selection of the training settings} on 1,133 held-out V3Det photos. \apm is measured on the 400
  person-free photos and \vpr on the 300 person-present photos (\%). Differences to the baseline model are given in
  points.}
  \label{tab:supp_sweep}
  \resizebox{\linewidth}{!}{\begin{tabular}{l cc rr rr c}
\toprule
model & $\beta$ & lr & \apm & $\Delta$\apm & \vpr & $\Delta$\vpr & passes guards \\
\midrule
MiniCPM-V-4.7-1B & 0.1 & 1e{-}5 & 4.5 & -7.2 & 90.7 & -0.7 & yes \\
MiniCPM-V-4.7-1B & 0.1 & 5e{-}6 & 5.5 & -6.2 & 92.0 & +0.7 & yes \\
MiniCPM-V-4.7-1B & 0.3 & 1e{-}5 & 5.8 & -6.0 & 91.3 & +0.0 & yes \\
MiniCPM-V-4.7-1B & 0.3 & 5e{-}6 & 8.8 & -3.0 & 91.0 & -0.3 & yes \\
Qwen3.5-9B & 0.1 & 5e{-}6 & 4.8 & -16.0 & 96.0 & -1.7 & yes \\
Qwen3.5-9B & 0.1 & 1e{-}5 & 5.0 & -15.7 & 93.0 & -4.7 & no \\
Qwen3.5-9B & 0.3 & 5e{-}6 & 6.0 & -14.8 & 94.3 & -3.3 & no \\
\bottomrule
\end{tabular}
}
\end{table}

\section{Results for all 22 VLMs}
\label{sec:supp_general}

\Cref{tab:generalization} reports every measure for the 22 aligned VLMs. On average, alignment reduces \apm by 80\%
in relative terms for the curators and by 76\% for the VLMs with no role in \dsname. Two models improve less than the
others. The first is Qwen3.5-4B, whose caption format changes under the shared training settings. Only 34\% of its
aligned captions consist of one sentence, and its captions grow from 26 to 63 words. Its first sentences mention an
absent person in only 5.9\% of the images, which means that its remaining attributions lie in the added sentences. The
second is InternVL3.5-14B, which keeps its format (19 words) and shows a genuinely partial effect. Idefics3-8B, the
model that starts closest to the human captions, still halves its rate.

The results are stable across seeds. Three seeds of the full method give an \apm of 8.4\%, 9.0\% and 9.6\% for
MiniCPM-V-4.7-1B and of 10.1\%, 9.8\% and 9.9\% for Qwen3.5-9B. The held-out V3Det photos show the same pattern for all
22 models. The caption length changes by a median of $+8\%$. The largest changes are those of Mistral-Small-3.1-24B,
whose captions become 73\% longer but still consist of one sentence, and of Qwen3.8-27B, whose captions become 61\%
shorter because the aligned model no longer ignores the one-sentence instruction. On the public preference sets, all
22 models mention people less often on the images for which human judges preferred a response without a person (set
A, from 37.0\% to 18.9\%), and they keep mentioning people on the images for which the judges preferred a response
with one (set B, from 89.4\% to 85.8\%).

\begin{table*}[t]
  \centering
  \small
  \setlength{\tabcolsep}{3.5pt}
  \caption{\textbf{All 22 VLMs} before and after alignment on \dsname with the shared training settings. $\Delta$\apm
  is given with its Holm-corrected scene-bootstrap $p$-value, and the asserted \apm is measured after alignment. Set A
  reports the person mentions on public images for which human judges preferred the response without a person.}
  \label{tab:generalization}
  \resizebox{\textwidth}{!}{\begin{tabular}{@{}l r rrl r rr rr rr rr rr@{}}
\toprule
 & & \multicolumn{3}{c}{\apm\,$\downarrow$} & asserted & \multicolumn{2}{c}{\pama\,$\uparrow$} & \multicolumn{2}{c}{\ppma\,$\uparrow$} & \multicolumn{2}{c}{\vpr} & \multicolumn{2}{c}{\ehw\,$\uparrow$} & \multicolumn{2}{c}{set A\,$\downarrow$} \\
\cmidrule(lr){3-5}\cmidrule(lr){7-8}\cmidrule(lr){9-10}\cmidrule(lr){11-12}\cmidrule(lr){13-14}\cmidrule(lr){15-16}
VLM & params & base & ours & $\Delta$ ($p$) & ours & base & ours & base & ours & base & ours & base & ours & base & ours \\
\midrule
\multicolumn{16}{@{}l}{\emph{data curators (8)}} \\
Gemma-4-12B & 12B & 42.9 & 5.6 & $-$37.3 ($<$0.05) & 2.7 & 59.1 & 87.8 & 95.5 & 94.6 & 96.7 & 94.8 & 41.3 & 43.7 & 33.1 & 17.9 \\
Gemma-4-31B & 31B & 56.9 & 9.8 & $-$47.1 ($<$0.05) & 2.7 & 48.4 & 85.5 & 94.8 & 95.1 & 96.0 & 95.7 & 40.3 & 43.4 & 38.1 & 19.1 \\
MiniCPM-V-4.7-1B & 1.55B & 43.5 & 8.4 & $-$35.1 ($<$0.05) & 2.9 & 58.9 & 85.8 & 94.3 & 94.7 & 94.9 & 95.2 & 41.2 & 43.6 & 34.6 & 21.5 \\
MiniCPM-V-4.7-35B-A3B & 35.2B & 46.5 & 15.4 & $-$31.2 ($<$0.05) & 4.4 & 58.3 & 82.0 & 95.0 & 94.7 & 96.7 & 95.7 & 41.4 & 43.4 & 35.6 & 23.1 \\
Muse-Glimmer-30B & 29.6B & 45.2 & 14.3 & $-$30.9 ($<$0.05) & 4.7 & 58.6 & 82.0 & 92.1 & 93.0 & 93.8 & 93.5 & 41.0 & 43.0 & 35.5 & 22.3 \\
Qwen3-VL-8B & 8B & 59.0 & 8.9 & $-$50.2 ($<$0.05) & 3.3 & 46.6 & 86.3 & 94.6 & 94.0 & 96.8 & 94.3 & 40.2 & 43.5 & 47.7 & 20.3 \\
Qwen3.5-9B & 9B & 64.8 & 10.1 & $-$54.7 ($<$0.05) & 3.5 & 42.3 & 85.1 & 94.7 & 94.8 & 96.3 & 95.4 & 40.4 & 43.7 & 47.3 & 18.9 \\
Qwen3.8-27B & 27B & 65.2 & 12.0 & $-$53.2 ($<$0.05) & 2.9 & 42.8 & 83.7 & 94.3 & 95.0 & 97.2 & 95.7 & 40.5 & 43.4 & 38.3 & 17.3 \\
\midrule
\multicolumn{16}{@{}l}{\emph{no role (14)}} \\
Idefics3-8B & 8.5B & 18.1 & 8.7 & $-$9.3 ($<$0.05) & 3.5 & 78.2 & 85.3 & 93.3 & 94.2 & 92.8 & 94.1 & 42.8 & 43.6 & 20.1 & 13.9 \\
SmolVLM2-2.2B & 2.2B & 39.5 & 6.6 & $-$32.8 ($<$0.05) & 2.9 & 62.6 & 87.3 & 95.4 & 95.7 & 97.5 & 96.6 & 41.7 & 43.8 & 26.9 & 14.4 \\
InternVL3.5-1B & 1.1B & 60.4 & 8.7 & $-$51.7 ($<$0.05) & 4.5 & 45.4 & 86.2 & 95.0 & 95.5 & 97.8 & 96.3 & 40.6 & 43.8 & 45.9 & 18.0 \\
InternVL3.5-14B & 15.1B & 39.5 & 20.8 & $-$18.7 ($<$0.05) & 3.5 & 63.9 & 78.0 & 95.3 & 95.1 & 97.3 & 96.3 & 41.6 & 42.7 & 33.4 & 21.9 \\
LLaVA-OneVision-7B & 8B & 45.5 & 8.0 & $-$37.5 ($<$0.05) & 3.2 & 57.7 & 86.7 & 94.9 & 95.2 & 97.7 & 95.8 & 41.3 & 43.8 & 29.2 & 15.9 \\
Mistral-Small-3.1-24B & 24B & 55.3 & 10.5 & $-$44.7 ($<$0.05) & 3.9 & 49.9 & 84.8 & 94.6 & 94.6 & 97.0 & 95.4 & 40.1 & 43.7 & 41.5 & 18.9 \\
Qwen2.5-VL-3B & 3.8B & 29.5 & 7.5 & $-$22.0 ($<$0.05) & 3.5 & 70.2 & 86.7 & 94.9 & 93.8 & 95.5 & 93.4 & 42.3 & 43.8 & 16.9 & 12.6 \\
Qwen2.5-VL-7B & 8.3B & 34.2 & 6.5 & $-$27.7 ($<$0.05) & 3.0 & 67.1 & 87.6 & 95.1 & 95.1 & 96.6 & 95.9 & 41.9 & 43.9 & 28.6 & 17.3 \\
Qwen3-VL-2B & 2.1B & 46.5 & 7.1 & $-$39.5 ($<$0.05) & 2.6 & 57.7 & 87.3 & 94.6 & 94.3 & 96.0 & 94.4 & 41.3 & 43.8 & 36.4 & 16.8 \\
Qwen3-VL-4B & 4.4B & 63.6 & 12.2 & $-$51.4 ($<$0.05) & 3.6 & 43.3 & 84.0 & 94.6 & 94.7 & 96.8 & 96.1 & 40.2 & 43.3 & 52.1 & 23.1 \\
Qwen3.5-0.8B & 0.8B & 55.3 & 6.6 & $-$48.6 ($<$0.05) & 4.1 & 50.4 & 86.8 & 95.2 & 94.5 & 96.9 & 94.3 & 40.9 & 43.9 & 44.8 & 17.7 \\
Qwen3.5-2B & 2B & 50.8 & 7.8 & $-$42.9 ($<$0.05) & 3.6 & 53.9 & 86.1 & 94.7 & 94.3 & 96.4 & 94.9 & 41.2 & 43.9 & 47.0 & 20.2 \\
Qwen3.5-4B & 4B & 59.8 & 24.2 & $-$35.5 ($<$0.05) & 3.0 & 45.8 & 73.6 & 94.8 & 94.6 & 96.8 & 96.6 & 40.7 & 43.1 & 44.4 & 26.0 \\
Qwen3.5-27B & 27B & 51.2 & 10.1 & $-$41.1 ($<$0.05) & 3.3 & 54.4 & 85.3 & 94.2 & 94.2 & 96.7 & 94.7 & 41.1 & 43.6 & 37.5 & 19.9 \\
\midrule
Human captions & & \multicolumn{2}{c}{12.5} & & 6.5 & \multicolumn{2}{c}{82.8} & \multicolumn{2}{c}{93.1} & \multicolumn{2}{c}{93.1} & \multicolumn{2}{c}{43.0} & \multicolumn{2}{c}{--} \\
\bottomrule
\end{tabular}
}
\end{table*}

\section{Details of the comparison with public preference datasets}
\label{sec:supp_comparison}

\paragraph{Public preference datasets used for training.} We convert each dataset to (image, prompt, chosen, rejected)
pairs. After removing the images that occur in the public preference sets described below, we draw a fixed random
sample of 5,844 pairs, except for RLHF-V, which has only 5,415 usable pairs. For VLFeedback, we pair the highest-rated
with the lowest-rated response and use two of its nine shards. For CSR we use its first iteration, and for HalluVL-DPO
its mixture subset. We dropped HA-DPO~\cite{zhao2023hadpo}, because its Visual Genome images could not be downloaded
and its release contains none. The mDPO row uses our pairs with the mDPO objective~\cite{wang2024mdpo}. We train every
row on Qwen3.5-9B with our training settings. The independent judge of caption quality is Qwen3.5-27B, which has no
role in the data. We ask it the question of the preference study in both orders and count a win only when both orders
agree.

\paragraph{Public preference sets used for evaluation.} We build two evaluation sets from four preference datasets
whose choices were made by people: Polaris~\cite{wada2024polos}, MM-RLHF~\cite{zhang2025mmrlhf},
LLaVA-Human-Preference~\cite{sun2024llavarlhf} and RLHF-V~\cite{yu2024rlhfv}. Of their 18,357 candidate pairs, we take
the 4,951 in which exactly one response mentions a person, where the lexicon and the classifier must agree on both
responses. Set A contains the pairs in which the judges chose the response without a person, and set B the pairs in
which they chose the response with a person. We keep one pair per image, exclude every image of our training data,
and match A and B within each dataset on the length of the chosen response. This yields 1,813 images per set. The
evaluated model captions each image with the \css prompt.

\section{Ablations results}
\label{sec:supp_ablations}

\textbf{Ablations on training: }\Cref{tab:supp_ablations} lists every variant that we trained. A 50/50
mix of person-absent and person-present pairs is weaker than our 75/25 mix. With a quarter of the data, Qwen3.5-9B
keeps more than half of its attributions, whereas with half of the data both models stay close to the full method.
Supervised fine-tuning on the chosen captions is weaker than DPO for Qwen3.5-9B. The IPO~\cite{azar2024ipo} and
RPO~\cite{pang2024rpo} losses trade a lower \apm for a lower \ppma. A larger $\beta$ of 0.3 is weaker, and a LoRA rank
of 64 is on par with rank 16. Labelling the person-present pairs by rule rather than by the judge is weaker for
MiniCPM-V-4.7-1B. Finally, three seeds of the full method give an \apm of 8.4\%, 9.0\% and 9.6\% for MiniCPM-V-4.7-1B
and of 10.1\%, 9.8\% and 9.9\% for Qwen3.5-9B.

\begin{table*}[t]
  \centering
  \small
  \caption{\textbf{All ablations} on MiniCPM-V-4.7-1B and Qwen3.5-9B (\benchname measures and \pca, \%).}
  \label{tab:supp_ablations}
  \resizebox{\textwidth}{!}{\begin{tabular}{@{}l rrrrr rrrrr @{}}
\toprule
 & \multicolumn{5}{c}{MiniCPM-V-4.7-1B} & \multicolumn{5}{c}{Qwen3.5-9B} \\
\cmidrule(lr){2-6}\cmidrule(lr){7-11}
Variant & APM$\downarrow$ & PAMA$\uparrow$ & PPMA$\uparrow$ & EHW$\uparrow$ & PJA$\uparrow$ & APM$\downarrow$ & PAMA$\uparrow$ & PPMA$\uparrow$ & EHW$\uparrow$ & PJA$\uparrow$ \\
\midrule
base model & 43.5 & 58.9 & 94.3 & 41.2 & 76.9 & 64.8 & 42.3 & 94.7 & 40.4 & 72.3 \\
\midrule
\dsname (full recipe) & 8.4 & 85.8 & 94.7 & 43.6 & 82.5 & 10.1 & 85.1 & 94.8 & 43.7 & 83.6 \\
\midrule
own curator's data only & 38.1 & 63.4 & 94.8 & 41.6 & 78.9 & 13.1 & 83.2 & 94.6 & 43.6 & 82.1 \\
own data, scene judge prompt & -- & -- & -- & -- & -- & 19.7 & 78.3 & 94.8 & 43.2 & 81.0 \\
judge: one order only & 8.0 & 85.7 & 94.8 & 43.6 & 83.0 & 10.1 & 85.2 & 94.5 & 43.7 & 83.0 \\
no judge (rule labels) & 13.3 & 82.8 & 95.2 & 43.3 & 81.9 & 13.1 & 83.3 & 94.9 & 43.5 & 82.9 \\
present pairs: rule labels & 13.9 & 82.4 & 95.1 & 43.2 & 81.2 & 11.4 & 84.5 & 94.8 & 43.6 & 83.8 \\
no person-present pairs & 5.0 & 87.0 & 90.7 & 43.6 & 82.1 & 5.9 & 87.9 & 94.2 & 44.0 & 83.5 \\
50 / 50 absent / present & 11.3 & 84.6 & 95.0 & 43.5 & 82.7 & 13.3 & 83.6 & 94.4 & 43.6 & 82.8 \\
25\% of the data & 10.2 & 84.7 & 94.5 & 43.5 & 82.6 & 39.0 & 63.3 & 94.8 & 42.0 & 78.3 \\
50\% of the data & 8.0 & 86.4 & 94.5 & 43.6 & 83.3 & 11.9 & 84.3 & 94.8 & 43.7 & 82.5 \\
no length balancing & 11.6 & 83.5 & 94.7 & 43.5 & 81.9 & 13.1 & 83.0 & 94.6 & 43.6 & 81.3 \\
IPO loss & 6.6 & 86.2 & 94.0 & 43.9 & 83.3 & 6.2 & 86.9 & 94.1 & 43.9 & 84.1 \\
RPO loss & 5.0 & 88.1 & 94.0 & 43.7 & 83.9 & 8.9 & 86.2 & 94.9 & 43.5 & 83.6 \\
mDPO loss & 7.7 & 86.5 & 94.5 & 43.6 & 83.0 & 11.3 & 84.6 & 94.7 & 43.4 & 82.5 \\
$\beta$ = 0.3 & 11.9 & 84.1 & 94.6 & 43.3 & 82.2 & 14.9 & 82.2 & 94.8 & 43.4 & 82.4 \\
LoRA rank 64 & 8.7 & 86.1 & 94.6 & 43.9 & 82.6 & 9.5 & 85.4 & 94.7 & 43.6 & 84.1 \\
SFT on chosen & 9.5 & 85.1 & 94.2 & 43.4 & 83.6 & 15.4 & 81.8 & 95.3 & 43.1 & 83.5 \\
shuffled labels & 43.2 & 58.7 & 94.2 & 41.2 & 76.7 & 67.6 & 39.9 & 94.9 & 40.2 & 72.1 \\
\midrule
human captions & 12.5 & 82.8 & 93.1 & 43.0 & -- & 12.5 & 82.8 & 93.1 & 43.0 & -- \\
\bottomrule
\end{tabular}
}
\end{table*}



\begin{table}[t]
  \centering
  \small
  \setlength{\tabcolsep}{3.5pt}
  \caption{\textbf{Ablations} on two curators (\%). Each row makes one change relative to the full method. All
  variants and measures are in \cref{tab:supp_ablations}.}
  \label{tab:ablations}
  \resizebox{\linewidth}{!}{\begin{tabular}{@{}l rrr@{\hspace{5pt}}|@{\hspace{5pt}}rrr@{}}
\toprule
 & \multicolumn{3}{c@{\hspace{5pt}}|@{\hspace{5pt}}}{MiniCPM-V-4.7-1B} & \multicolumn{3}{c}{Qwen3.5-9B} \\
\cmidrule(lr){2-4}\cmidrule(lr){5-7}
Variant & \apm$\downarrow$ & \pama$\uparrow$ & \ppma$\uparrow$ & \apm$\downarrow$ & \pama$\uparrow$ & \ppma$\uparrow$ \\
\midrule
Baseline model & 43.5 & 58.9 & 94.3 & 64.8 & 42.3 & 94.7 \\
\midrule
\textbf{\textcolor{nbA}{+}\textcolor{nbA}{N}\textcolor{nbB}{O}\textcolor{nbC}{B}\textcolor{nbD}{O}\textcolor{nbE}{D}\textcolor{nbF}{Y}} (full method) & 8.4 & 85.8 & 94.7 & 10.1 & 85.1 & 94.8 \\
\midrule
Own curator's pairs only & 38.1 & 63.4 & 94.8 & 13.1 & 83.2 & 94.6 \\
No judge (rewrite always chosen) & 13.3 & 82.8 & 95.2 & 13.1 & 83.3 & 94.9 \\
Shuffled labels & 43.2 & 58.7 & 94.2 & 67.6 & 39.9 & 94.9 \\
No person-present pairs & 5.0 & 87.0 & 90.7 & 5.9 & 87.9 & 94.2 \\
No length balancing & 11.6 & 83.5 & 94.7 & 13.1 & 83.0 & 94.6 \\
25\% of the pairs & 10.2 & 84.7 & 94.5 & 39.0 & 63.3 & 94.8 \\
SFT on chosen captions & 9.5 & 85.1 & 94.2 & 15.4 & 81.8 & 95.3 \\
\midrule
\textit{Human captions} & 12.5 & 82.8 & 93.1 & 12.5 & 82.8 & 93.1 \\
\bottomrule
\end{tabular}
}
\end{table}

\textbf{Ablations on NOBODY:} We vary one component of our approach at a time on MiniCPM-V-4.7-1B and Qwen3.5-9B. Results are shown in (\cref{tab:ablations}). 


\textbf{Ablations on judge prompt: }In \dsname, the judge is asked the question of the human study. To test whether self-judging depends on the wording
of this question, we repeated the whole pipeline on Qwen3.5-9B with a second judge prompt. This \emph{scene prompt} is
phrased like the captioning task and never mentions people. It is given verbatim in \cref{sec:supp_prompts}, where the
missing space after ``scene.'' is part of the original string. We refer to the default prompt as the \emph{study
question}.

The first finding is that the penalty for an attributed person remains. On the items of the human study, the judge with the scene
prompt chooses the VLM caption in 43.4\% of the pairs in which only this caption mentions a person, and in 62.0\% of
the pairs in which neither caption does. The resulting odds ratio of 0.47 (95\% CI [0.35, 0.62]; one-sided Fisher's
test, $p=8\cdot10^{-7}$) is close to that of the human judges on the same items (0.43 [0.35, 0.53]). With the study
question, the same VLM penalizes attributions more strongly than the human judges do (0.21). The scene prompt favors
the VLM caption more often overall, and its verdicts are consistent less often (71\% \vs 85\%), but the penalty for
an added person remains (\cref{tab:judge_prompt}a).

We judged the identical 2,645 person-absent
and 2,481 person-present pairs with both prompts. On the pairs for which both prompts give a consistent verdict, the
two prompts agree in 98.5\% and 99.0\% of the cases, respectively. However, the scene prompt gives a consistent verdict
less often (79.6\% \vs 84.9\% on person-absent pairs), so fewer pairs survive. After the same cleaning and balancing,
986 pairs remain (740 person-absent and 246 person-present), compared with 1,180 for the study question.

We trained Qwen3.5-9B through DPO on the scene-prompt pairs, with the
shared training settings and two epochs, as in every run that uses a curator's own data only. The aligned model drops
most attributions and keeps mentioning visible people (\cref{tab:judge_prompt}b). On the held-out V3Det photos, its
\apm falls from 20.8\% to 5.8\%, and its caption length is unchanged.

Compared with the model trained on the full set of study-question pairs, the
model trained on the scene-prompt pairs mentions people in 6.6 points more of the person-free scenes (95\% CI [4.0,
9.3]; McNemar's test, $p<10^{-4}$). We therefore add a size-matched control, in which the study-question pairs are
subsampled to the same 740 person-absent and 246 person-present pairs. Compared with this control, the difference is
only $+1.5$ points (95\% CI [$-0.7$, $+3.7$], $p=0.12$), and \ehw is identical.


\begin{table*}[t]
\centering
\small
\setlength{\tabcolsep}{4pt}
\caption{Judge-prompt ablation on Qwen3.5-9B. (a) The judge's penalty for a caption that adds a person, on the human-study items (CSS images without people; odds ratio of choosing the model caption when only it mentions a person vs.\ when neither caption does; both presentation orders must agree; 95\% bootstrap CI). (b) DPO on the VLM's own caption pairs judged with each prompt (same images, captions, cleaning, length balancing and recipe; the size-matched row is a random subsample of the study-question pairs with the scene-prompt dataset's composition), evaluated on the CSS benchmark (APM, asserted APM, VPM, EHW, in \%) and the public A/B sets (person-mention rate).}
\label{tab:judge_prompt}
\begin{tabular}{lccc}
\toprule
(a) judge & prompt & OR (95\% CI) & order consistency \\
\midrule
humans & study question & 0.43 [0.35, 0.53] & -- \\
Qwen3.5-9B & study question & 0.21 [0.14, 0.29] & 85\% \\
Qwen3.5-9B & scene prompt & 0.47 [0.35, 0.62] & 71\% \\
\bottomrule
\end{tabular}
\par\medskip
\begin{tabular}{lrcccccc}
\toprule
(b) training data & pairs & APM$\downarrow$ & asserted$\downarrow$ & VPM$\uparrow$ & EHW$\uparrow$ & set A$\downarrow$ & set B \\
\midrule
base model & -- & 64.8 & 42.8 & 96.3 & 40.4 & 47.3 & 91.9 \\
own pairs, study question & 1,180 & 13.1 & 3.2 & 95.7 & 43.6 & 23.0 & 87.1 \\
own pairs, study question, size-matched & 986 & 18.2 & 4.7 & 96.0 & 43.2 & 25.4 & 88.3 \\
own pairs, scene prompt & 986 & 19.7 & 5.3 & 96.3 & 43.2 & 26.2 & 88.2 \\
NOBODY (8 curators, study question) & 5,844 & 10.1 & 3.5 & 95.4 & 43.7 & 18.9 & 86.0 \\
\midrule
humans & & 12.5 & 6.5 & 93.1 & -- & -- & -- \\
\bottomrule
\end{tabular}
\end{table*}

\section{VLM general abilities after alignment}
\label{sec:supp_side}

\Cref{tab:side_effects} reports four benchmarks for the eight curators before and after alignment on \dsname, together
with the caption-quality win rate of the independent judge. For POPE~\cite{li2023pope}, we report the F1 score over
500 questions in each of its three splits. For CHAIR$_i$~\cite{rohrbach2018object}, the models describe 500 COCO
validation images with the prompt ``Describe this image in detail.'' For MMHal-Bench~\cite{sun2024llavarlhf}, we
report the share of the 96 answers that phi-4 rates below 3 with the rating template of the benchmark, so that one
answer corresponds to about one point. For MMStar~\cite{chen2024mmstar}, we use its 1,500 questions. The differences
on POPE, CHAIR$_i$ and MMStar are about one point or less, and the changes on MMHal-Bench, which amount to one to four
answers in either direction, are within noise. The independent judge prefers the aligned captions for seven curators
and is indifferent for Qwen3.8-27B, whose baseline captions often break the one-sentence instruction. Averaged over
the curators, it prefers the aligned caption in 77\% of its consistent verdicts.

We also examined what replaces the attributions. Most captions that asserted a person become captions without any
person. Captions that only imply an agent (``ready to be enjoyed'', as labelled by the classifier of
\cref{sec:supp_detectors}) become more frequent in 16 of the 22 models ($+6.3$ points on average). In addition,
removed non-person objects are mentioned slightly more often (the persistence rises from 16.9\% to 19.8\%,
$p<10^{-3}$).

\begin{table*}[t]
  \centering
  \small
  \caption{\textbf{Side effects} on hallucination and general benchmarks (\%), for the baseline model (base) and
  after alignment on \dsname (ours). The win rate is the share of consistent verdicts of the independent judge
  Qwen3.5-27B (both orders, \css images) that prefer the aligned caption over the baseline caption.}
  \label{tab:side_effects}
  \begin{tabular}{@{}l rr rr rr rr r@{}}
\toprule
 & \multicolumn{2}{c}{POPE F1$\uparrow$} & \multicolumn{2}{c}{CHAIR$_i\downarrow$} & \multicolumn{2}{c}{MMHal hal.$\downarrow$} & \multicolumn{2}{c}{MMStar$\uparrow$} & win rate \\
Data curator & base & ours & base & ours & base & ours & base & ours & of ours \\
\midrule
Gemma-4-12B & 85.7 & 85.7 & 16.7 & 17.0 & 9.4 & 12.5 & 44.3 & 44.3 & 88.9 \\
Gemma-4-31B & 89.1 & 89.2 & 15.5 & 16.4 & 13.5 & 11.5 & 47.8 & 47.2 & 95.0 \\
MiniCPM-V-4.7-1B & 87.4 & 87.6 & 28.0 & 28.3 & 6.2 & 8.3 & 36.3 & 35.9 & 85.5 \\
MiniCPM-V-4.7-35B-A3B & 88.9 & 88.6 & 28.5 & 28.5 & 6.2 & 8.3 & 55.3 & 55.4 & 73.2 \\
Muse-Glimmer-30B & 87.1 & 86.8 & 29.3 & 29.5 & 3.1 & 4.2 & 74.3 & 73.5 & 78.8 \\
Qwen3-VL-8B & 88.8 & 88.5 & 25.3 & 24.6 & 3.1 & 7.3 & 61.2 & 61.6 & 87.3 \\
Qwen3.5-9B & 88.7 & 88.9 & 29.0 & 28.6 & 3.1 & 5.2 & 52.5 & 52.5 & 59.7 \\
Qwen3.8-27B & 88.9 & 90.1 & 30.7 & 29.8 & 9.4 & 8.3 & 50.9 & 50.2 & 48.1 \\
\bottomrule
\end{tabular}

\end{table*}

\section{Other description prompts}
\label{sec:supp_other_prompts}

\begin{table}[t]
  \centering
  \small
  \caption{\textbf{Baseline Qwen3.5-9B under three prompts} on the \css images. \apm and \vpr are the shares of
  captions (\%) that have person attribution on person-absent and person-present images respectively.}
  \label{tab:supp_other_prompts}
  \begin{tabular}{@{}l rr@{}}
\toprule
Prompt & \apm & \vpr \\
\midrule
\css prompt & 64.8 & 96.3 \\
``Describe this image.'' & 58.1 & 98.9 \\
``Describe this image in detail.'' & 66.3 & 99.3 \\
\midrule
Human captions (\css) & 12.5 & 93.1 \\
\bottomrule
\end{tabular}

\end{table}

Unwarranted person attribution is not specific to the \css prompt. We caption the \css images with the baseline
Qwen3.5-9B under two open description prompts, with the settings of \cref{sec:supp_impl} and a limit of 768 tokens
(\cref{tab:supp_other_prompts}). Under both prompts, the model mentions a person in more than half of its
descriptions of person-absent images, as it does under the \css prompt. 

\section{Large, central and salient objects}
\label{sec:supp_objects}

\begin{table*}[t]
  \centering
  \small
  \renewcommand{\arraystretch}{1.15}
  \caption{\textbf{Examples} of baseline and aligned captions on \css images. The captions are verbatim, and ``\dots''
  marks a cut.}
  \label{tab:supp_examples}
  \begin{tabular}{@{}p{0.13\textwidth}p{0.12\textwidth}p{0.31\textwidth}p{0.35\textwidth}@{}}
  \toprule
  model & case & baseline caption & after alignment on \dsname \\
  \midrule
  \csname @@input\endcsname tables/tab_examples.tex
  \bottomrule
  \end{tabular}
\end{table*}

The \css analyses also concern the largest, the most central and the most salient object of a scene. We therefore
applied the same self-judged method to these objects, using Qwen3-VL-8B on V3Det photos with a clear target object.
For each photo, the curator's regular caption is paired with its rewrite under the instruction ``Do not mention the
\{label\}.'' To detect whether a caption mentions the target, we use an evidence-based check, in which the curator
quotes the words that refer to the outlined object and the quote is verified against the caption, together with
reference lists of names for each target.

This method yields almost no data, for two reasons. First, when the curator is told not to mention the main object,
it still refers to it in 83\% of its rewrites on average (\eg ``a small bird'' for a named bird species), because the
scene cannot be described without the object. Second, when the curator does omit the object, it prefers, as a judge,
the caption with the object in 88\% of its consistent verdicts. Of the 18,000 candidate photos for the largest
object, only 244 pairs remain. For the most central object, 85 pairs remain of 8,641 photos, and for the most salient
object, 48 of 3,441. This outcome matches the human judges. In the preference model of \cref{sec:supp_bench},
mentioning the largest or the most central object has an odds ratio of 0.99 and 1.04, respectively, \ie the judges do
not penalize these mentions. Note that these quantities are mention rates and not the correlation-based sensitivities
of the \css paper. They indicate where self-judged preferences do not apply, rather than a failure that should be
fixed.

\section{Additional examples}
\label{sec:supp_examples}

\Cref{tab:supp_examples} shows examples of baseline and aligned captions, which we selected with deterministic rules.
\emph{Fixes} are images without a visible person for which the baseline caption asserts a person, the aligned caption
mentions none and no human caption mentions one. \emph{Kept} cases are images with a visible person for which both
captions mention one. Failures are of two kinds: images without people for which the aligned caption still mentions a
person although no human caption does, and images with a visible person whom the aligned caption no longer mentions.
Many of the remaining mentions are implied or hypothetical (``suggesting someone might be working''), and some of the
apparent omissions are misses of the detector (``a silhouetted beachgoer'').

\section{Prompts}
\label{sec:supp_prompts}

We list all prompts verbatim. Curly braces mark placeholders, and a leading space belongs to the string. Every query
is sent in a new chat with greedy decoding.

\paragraph{Captioning.} We use the \css prompt for the captions of every model, in the curation, in \benchname, on the
public preference sets and on the V3Det photos:
\begin{lstlisting}
Make your best guess of what might be happening in this scene in one sentence. Avoid mentioning objects that do not aid in understanding the context of the scene.
\end{lstlisting}
In the curation, we append the following rewrite instruction to the \css prompt:
\begin{lstlisting}
 Do not mention any people.
\end{lstlisting}
For the prompting baseline of \cref{tab:comparison}, we append the following instruction to the \css prompt:
\begin{lstlisting}
 Do not mention people or animals unless they are visible in the image.
\end{lstlisting}
For the length-matched VLM captions of the preference studies, we append the following instruction to the \css
prompt, where $n$ is the number of words:
\begin{lstlisting}
 Your response must contain exactly {n} {word}.
\end{lstlisting}
For the CHAIR descriptions, we use the following prompt:
\begin{lstlisting}
Describe this image in detail.
\end{lstlisting}

\paragraph{Judge with the study question.} The curation and all VLM judges use the following system and user prompts:
\begin{lstlisting}
You are evaluating image captions. You will be shown an image and two candidate captions. Decide which caption better describes the image. Respond with only the single digit 1 or 2 and nothing else.
\end{lstlisting}
\begin{lstlisting}
Which caption better describes the image?

Caption 1: {caption1}
Caption 2: {caption2}

Answer with only 1 or 2.
\end{lstlisting}

\paragraph{Judge with the scene prompt.} The ablation of \cref{sec:supp_ablations} uses the following system and user
prompts:
\begin{lstlisting}
You are evaluating image captions. You will be shown an image and two candidate captions. Decide which caption better describes what might be happening in the scene. The better caption should avoid mentioning objects that do not aid in understanding the context of the scene.Respond with only the single digit 1 or 2 and nothing else.
\end{lstlisting}
\begin{lstlisting}
Which caption better describes what might be happening in the scene?

Caption 1: {caption1}
Caption 2: {caption2}

Answer with only 1 or 2.
\end{lstlisting}

\paragraph{Checks.} We use the first question below to check whether a person is visible (person-free V3Det photos
and public preference sets), and the second to check for a real person (person-present photos):
\begin{lstlisting}
Is there any person, or any part of a person (face, hand, arm, leg), visible anywhere in this image, including drawings, photos, posters, statues or reflections? Answer yes or no.
\end{lstlisting}
\begin{lstlisting}
Is there a real, clearly visible person in this image (not a drawing, photo, poster, statue, mannequin or toy)? Answer yes or no.
\end{lstlisting}
The person-mention classifier (phi-4, text only) uses the following system and user prompts:
\begin{lstlisting}
You classify image descriptions. Answer with exactly one word: asserted, implied or none.
\end{lstlisting}
\begin{lstlisting}
Description: "{caption}"

Does this description refer to a person (or people)?
- asserted: it states or presupposes that a person is present in the scene (e.g. "a man is cooking", "a child's toys are scattered while she plays", "someone is sitting on the bench").
- implied: a person is only mentioned as a generic, hypothetical, past or absent agent, an owner, or a purpose (e.g. "ready for guests", "waiting for someone to sit", "left by a child", "a kitchen where meals are prepared").
- none: no person is referred to at all.
Answer with one word: asserted, implied or none.
\end{lstlisting}

\paragraph{Objects.} The analysis of \cref{sec:supp_objects} uses, in this order, the rewrite instruction, the quote
check with the target outlined in red, the recall-oriented text check, and two checks that we abandoned during
development:
\begin{lstlisting}
 Do not mention the {label}.
\end{lstlisting}
\begin{lstlisting}
The red box marks one object in this image.
Description: "{caption}"
Which words of the description refer to the object inside the red box? Reply with the exact words copied from the description, or reply "none" if no words refer to it.
\end{lstlisting}
\begin{lstlisting}
Object: "{everyday}" ({label}{parent}).
Description: "{caption}"
Does the description refer to this object in any way: by its name, by a more general word for it, or by describing it without naming it? Answer yes or no.
\end{lstlisting}
\begin{lstlisting}
Here is a description of this image: "{caption}"
Does the description mention the object inside the red box? Answer yes or no.
\end{lstlisting}
\begin{lstlisting}
Description: "{caption}"
Does this description mention {label} or a synonym of it? Answer yes or no.
\end{lstlisting}

\paragraph{Prompts not used for any reported result.} For completeness, we also list three prompts that are defined
in our code but not used for any reported result: a faithfulness judge of an earlier pilot, a ``visible only''
suffix and a short captioning prompt.
\begin{lstlisting}
You are checking image descriptions for faithfulness. A good description makes a sensible guess about what is happening, but only mentions people, animals and objects that are actually visible in the image; it does not invent people or events that cannot be seen.

Description A: "{a}"
Description B: "{b}"

Which description is better? Answer with A or B.
\end{lstlisting}
\begin{lstlisting}
 Base your guess only on what is actually visible in the image.
\end{lstlisting}
\begin{lstlisting}
Describe this image in one sentence.
\end{lstlisting}

\end{document}